\pdfoutput=1
\documentclass[11pt]{article}

\usepackage[final]{acl}

\usepackage{times}
\usepackage{latexsym}

\usepackage[T1]{fontenc}
\usepackage[utf8]{inputenc}

\usepackage{microtype}

\usepackage{inconsolata}

\usepackage{graphicx}
\usepackage{times}
\usepackage{latexsym}
\usepackage{inconsolata}
\usepackage[utf8]{inputenc}
\usepackage{microtype}
\usepackage{inconsolata}
\usepackage{graphicx}
\usepackage{tcolorbox}
\usepackage{inconsolata}
\usepackage{microtype}
\usepackage{enumitem}
\usepackage{geometry}
\usepackage{amssymb}
\usepackage{mathrsfs}
\usepackage{amsmath}
\usepackage{hhline}
\usepackage{amsfonts}
\usepackage{multirow}
\usepackage{tabularx}
\usepackage{array}
\usepackage{booktabs}
\usepackage{verbdef}
\usepackage{colortbl}
\usepackage{color}
\usepackage{subcaption}
\usepackage{threeparttable}
\usepackage{float}
\usepackage{caption}
\usepackage{arydshln}
\usepackage{inconsolata}
\usepackage{rotating}
\usepackage{xcolor}
\usepackage{placeins}
\usepackage{makecell}
\usepackage{afterpage}
\usepackage{blindtext}
\usepackage{float}
\usepackage{pifont}
\usepackage{threeparttablex}
\usepackage[framemethod=TikZ]{mdframed}
\usepackage{hhline}
\usepackage{hyperref}
\definecolor{mygray}{gray}{0.9}
\definecolor{mypink}{rgb}{0.99,0.91,0.95}
\definecolor{Coral}{RGB}{245, 218, 210}
\definecolor{mycyan}{cmyk}{0.3,0,0,0}
\definecolor{Celadon}{RGB}{172, 225, 175}
\definecolor{Peach}{RGB}{255, 229, 180}
\definecolor{orange}{HTML}{FFF8E3}
\definecolor{green}{HTML}{DAE3D1}
\definecolor{blue}{HTML}{EAF8FF}
\newcommand{\PlanSum}{\texttt{\mbox{Plan-mPlug-Owl3}}\xspace}
\newcommand{\cmark}{\ding{51}}  
\newcommand{\xmark}{\ding{55}}
\title{What Is That Talk About? A Video-to-Text Summarization Dataset for Scientific Presentations}

\author{
 \textbf{Dongqi Liu\textsuperscript{$\Omega$}\thanks{This work was partially conducted during the research visit at the University of Edinburgh.}},
 \textbf{Chenxi Whitehouse\textsuperscript{$\Delta$}},
 \textbf{Xi Yu\textsuperscript{$\Omega$}},
 \textbf{Louis Mahon\textsuperscript{$\Theta$}},
 \textbf{Rohit Saxena\textsuperscript{$\Theta$}},
 \\
 \textbf{Zheng Zhao\textsuperscript{$\Theta$}},
 \textbf{Yifu Qiu\textsuperscript{$\Theta$}},
 \textbf{Mirella Lapata\textsuperscript{$\Theta$}},
 \textbf{Vera Demberg\textsuperscript{$\Omega$}\textsuperscript{$\Psi$}}
 \\
 \textsuperscript{$\Omega$}Saarland University, 
 \textsuperscript{$\Psi$}Max Planck Institute for Informatics \\
 \textsuperscript{$\Delta$}University of Cambridge, 
 \textsuperscript{$\Theta$}University of Edinburgh \\
 \small{
   \shortstack[c]{
     \textsuperscript{$\Omega$}\texttt{\{dongqi,xiyu,vera\}@lst.uni-saarland.de} \\
     \textsuperscript{$\Delta$}\texttt{chenxi.whitehouse@cl.cam.ac.uk} \\
     \textsuperscript{$\Theta$}\texttt{\{lmahon,rohit.saxena,zheng.zhao,yifu.qiu\}@ed.ac.uk}, \texttt{mlap@inf.ed.ac.uk}
   }
 }
}

\begin{document}
\maketitle
\begin{abstract}
Transforming recorded videos into concise and accurate textual summaries is a growing challenge in multimodal learning. This paper introduces VISTA, a dataset specifically designed for video-to-text summarization in scientific domains. VISTA contains 18,599 recorded AI conference presentations paired with their corresponding paper abstracts. We benchmark the performance of state-of-the-art large models and apply a plan-based framework to better capture the structured nature of abstracts. Both human and automated evaluations confirm that explicit planning enhances summary quality and factual consistency. However, a considerable gap remains between models and human performance, highlighting the challenges of our dataset. This study aims to pave the way for future research on scientific video-to-text summarization. The project information is available at \href{https://dongqi.me/projects/VISTA}{https://dongqi.me/projects/VISTA}.
\end{abstract}

\section{Introduction}

Large multimodal models (LMMs), which integrate components from different modalities through cross-modal alignment training \cite{koh2023grounding, cheng-etal-2023-edit, li2024seed, ahn-etal-2024-tuning, fu2025blink, wu2025comprehensive}, have achieved considerable progress in video-to-text summarization tasks for general-purpose content such as YouTube, movies, and news videos \cite{li-etal-2020-vmsmo, lin2023videoxum, krubinski-pecina-2023-mlask, hua2024v2xum, chen2024personalized, zhang2024multi, qiu2024mmsum, patil-etal-2024-refinesumm, mahon-lapata-2024-modular, mahon2024screenwriter}. However, many recent studies have highlighted that these LMMs exhibit reduced performance in scientific contexts, particularly when processing technical terminology and scientific visual elements like figures and tables \cite{li-etal-2024-multimodal-arxiv, lu2024mathvista, yue2024mmmu, bai2024hallucination, liang2024foundations, patil-etal-2024-refinesumm, huang2024pixels}. This performance gap might be largely attributed to the absence of specialized datasets for multimodal scientific content \cite{chen-etal-2024-m3av, hu2024mplugpaperowl, pramanick2024spiqa, zhang-etal-2024-comprehensive-survey}.

\begin{figure}[t]
  \centering \includegraphics[width=0.45\textwidth]{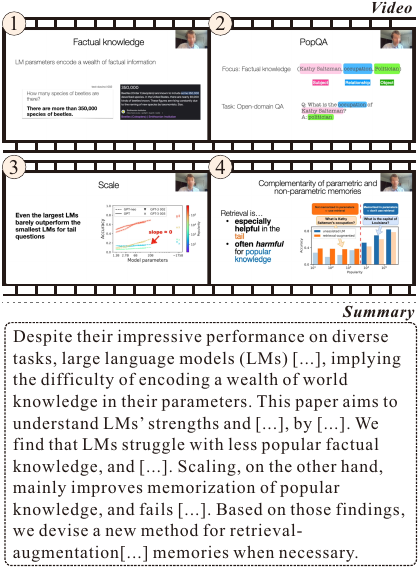}
  \caption{An example from VISTA: a conference presentation video (top) paired with the abstract of the corresponding paper (bottom). This data sample \cite{mallen-etal-2023-trust} was presented at ACL 2023 and received the Best Video Recordings award.}
  \label{fig:dataset_illustration}
\vspace{-15pt}
\end{figure}

Thus, we introduce \textbf{VISTA} (\textbf{\underline{Vi}}deo to \textbf{\underline{S}}cien\textbf{\underline{t}}ific \textbf{\underline{A}}bstract), an English dataset for video-to-text summarization in scientific domains. VISTA consists of 18,599 aligned pairs of conference presentation recordings and their corresponding paper abstracts, collected from leading conferences in computational linguistics (\href{https://aclanthology.org/}{ACL Anthology} including ACL, EMNLP, NAACL, EACL, Findings of *ACL) and machine learning (\href{https://icml.cc/}{ICML} and \href{https://neurips.cc/}{NeurIPS}). \autoref{fig:dataset_illustration} illustrates an example selected from VISTA.

We use the abstract of the paper as a proxy for the summary of the video and benchmark VISTA using several state-of-the-art (SOTA) large models, including closed-source LMMs (\texttt{Claude 3.5 Sonnet}, \texttt{Gemini 2.0}, \texttt{GPT-o1}), as well as video-specific open-source LMMs (\texttt{Video-LLaMA}, \texttt{Video-ChatGPT}, \texttt{mPLUG-Owl3}, etc.; \citealp[]{zhang-etal-2023-video, maaz-etal-2024-video, lin-etal-2024-video, ye2024mplug, li2024llava, li2025llama}). For comparison, we also include strong baselines: text-to-text model \texttt{LLaMA-3.1} \cite{touvron2023llama} and audio-to-text model \texttt{Qwen2-Audio} \cite{chu2024qwen2}. Experiments across zero-shot, QLoRA, and full fine-tuning settings reveal that in-domain fine-tuning improves summarization performance across different large models, and video-based models generally outperform text- and audio-based models on our dataset. However, end-to-end approaches may often struggle to capture the underlying structure of scientific abstracts \cite{liu2025explanatorysummarizationdiscoursedrivenplanning}. 

To address this, we explore a plan-based approach, which has been shown to improve coherence and factual grounding through a predefined planning component \cite{narayan-etal-2021-planning, narayan-etal-2023-conditional, liu2025explanatorysummarizationdiscoursedrivenplanning}. Unlike direct end-to-end generation, plan-based methods can leverage the fact that scientific abstracts often follow a well-defined format \cite{takeshita-etal-2024-aclsum}. By explicitly modeling the latent structure of the summary through a sequence of intermediate plans, the summary generation process can be better guided. Empirical results confirm that the plan-based method outperforms existing SOTA models in terms of summary quality and factual accuracy. This work also lays the groundwork for future investigations into the multimodal summarization of scientific videos.

\textbf{In summary, our contributions are as follows:}
\begin{itemize}[leftmargin=8pt,itemsep=1pt,topsep=1pt,parsep=1pt]
\item We present VISTA, a novel large-scale multimodal dataset with 18,599 video-summary pairs, tailored for summarizing scientific presentations from video recordings.
\item We establish benchmark performance on VISTA through a comprehensive evaluation of leading large (language/audio/multimodal) models.
\item We leverage a plan-based approach that consistently improves summary quality and factual accuracy over SOTA models.
\item We conduct error analysis, case studies, and human evaluations to identify the pivotal issues in the model-generated summaries.
\end{itemize}

\section{Related Work}
\paragraph{Video-to-Text Summarization} generates coherent summaries by integrating multimodal information \cite{hua2024v2xum}, supported by datasets like MSS \cite{li-etal-2017-multi}, VideoXum \cite{10334011}, MMSum \cite{qiu2024mmsum}, Hierarchical3D \cite{papalampidi-lapata-2023-hierarchical3d}, and \mbox{LfVS-T} \cite{argaw2024scaling}, spanning tasks from instructional videos to general web content \cite{li-etal-2017-multi, zhou2018towards, li2019video, li-etal-2020-vmsmo, liu-wan-2021-video, fu-etal-2021-mm, pu-etal-2022-two, krubinski-pecina-2023-mlask, han2023shot2story20k, he2023align, hua2024v2xum, islam2024video, qiu2024mmsum}. Technical advancements include hierarchical attention models \cite{sanabria2018how2}, extractive methods using multimodal features \cite{cho-etal-2021-streamhover, krubinski-pecina-2023-mlask}, and hybrid extractive-abstractive frameworks \cite{9939279, papalampidi-lapata-2023-hierarchical3d}. Transformer-based systems have further improved performance \cite{krubinski-pecina-2023-mlask,li-etal-2020-vmsmo,10.1145/3474085.3475321,mahon-lapata-2024-modular}. However, challenges in summarizing academic videos remain under-explored.

\paragraph{Scientific Text Summarization} condenses complex scholarly content into concise formats \cite{cachola-etal-2020-tldr, ju-etal-2021-leveraging-information, pu-etal-2023-incorporating, pu-demberg-2023-chatgpt}, supported by datasets like TalkSumm \cite{lev-etal-2019-talksumm} for academic video transcripts, SumSurvey \cite{liu-etal-2024-sumsurvey} for survey papers, ACLSum \cite{takeshita-etal-2024-aclsum} for ACL discourse, and SciNews \cite{pu-etal-2024-scinews} for simplifying research for broader audiences. M$^3$AV \cite{chen-etal-2024-m3av} supports tasks like ASR, TTS, and slide-script generation. Methods like RST-LoRA \cite{pu-demberg-2024-rst} and RSTformer \cite{pu-etal-2023-incorporating} improve discourse and structural summarization, while CiteSum \cite{mao-etal-2022-citesum} and SSR \cite{fatima-strube-2023-cross} focus on scalability and audience-specific customization. Despite these efforts, scientific summarization remains a challenging domain due to the inherent complexity and diversity of scholarly texts. 

\paragraph{Plan-based Summarization} employs structured representations to improve summary quality and reduce hallucinations \cite{narayan-etal-2021-planning, amplayo2021unsupervised, wang-etal-2022-guiding, narayan-etal-2023-conditional, liu2025explanatorysummarizationdiscoursedrivenplanning}. Research focuses on text-based planning with elements like entities \cite{narayan-etal-2021-planning, liu-chen-2021-controllable, huot-etal-2024-mplan}, keyword prompts \cite{creo2023prompting}, and question-answer pairs \cite{narayan-etal-2023-conditional}. Examples include PlanVerb \cite{canal2022planverb}, which converts task plans into natural language via semantic tagging, and domain-specific approaches that align with knowledge structures for improved quality \cite{srivastava-etal-2024-knowledge}. Blueprint-based frameworks utilize intermediate plans to create coherent narratives for visual storytelling \cite{liu-etal-2023-visual-storytelling}. However, plan-based strategies for multimodal tasks, particularly video-to-text summarization, have received limited attention. 

\section{The VISTA Dataset}

\paragraph{Data Acquisition and Cleaning}
VISTA is derived from computational linguistics and machine learning conferences, including \href{https://aclanthology.org/}{ACL Anthology} (ACL, EMNLP, NAACL, EACL, Findings of *ACL), \href{https://icml.cc/}{ICML}, and \href{https://neurips.cc/}{NeurIPS}, covering content from 2020 to 2024. All materials (paper abstracts and video recordings) are contributed by the respective paper authors, ensuring narrative consistency. Since these metadata are stored in XML/JSON files on their respective websites, no further data preprocessing (e.g., extracting abstracts from PDFs) is required. We collect paper titles, author lists, paper abstracts, links to papers, and presentation videos, in accordance with platform terms for academic research purposes (or obtain written confirmation).\footnote{We discuss copyright in \autoref{copyrights}.} To maintain one-to-one video-to-text alignments, we exclude samples that may cover multiple papers (e.g., tutorials, invited talks) and videos shorter than one minute or longer than 30 minutes.

\paragraph{Quality Control} We verify the data quality through both manual and automated checks. We discuss quality control guidelines and the results in Appendix \autoref{fig:quality_control} and \autoref{quality_control}, respectively.

\begin{figure}[t]
  \centering \includegraphics[width=0.49\textwidth]{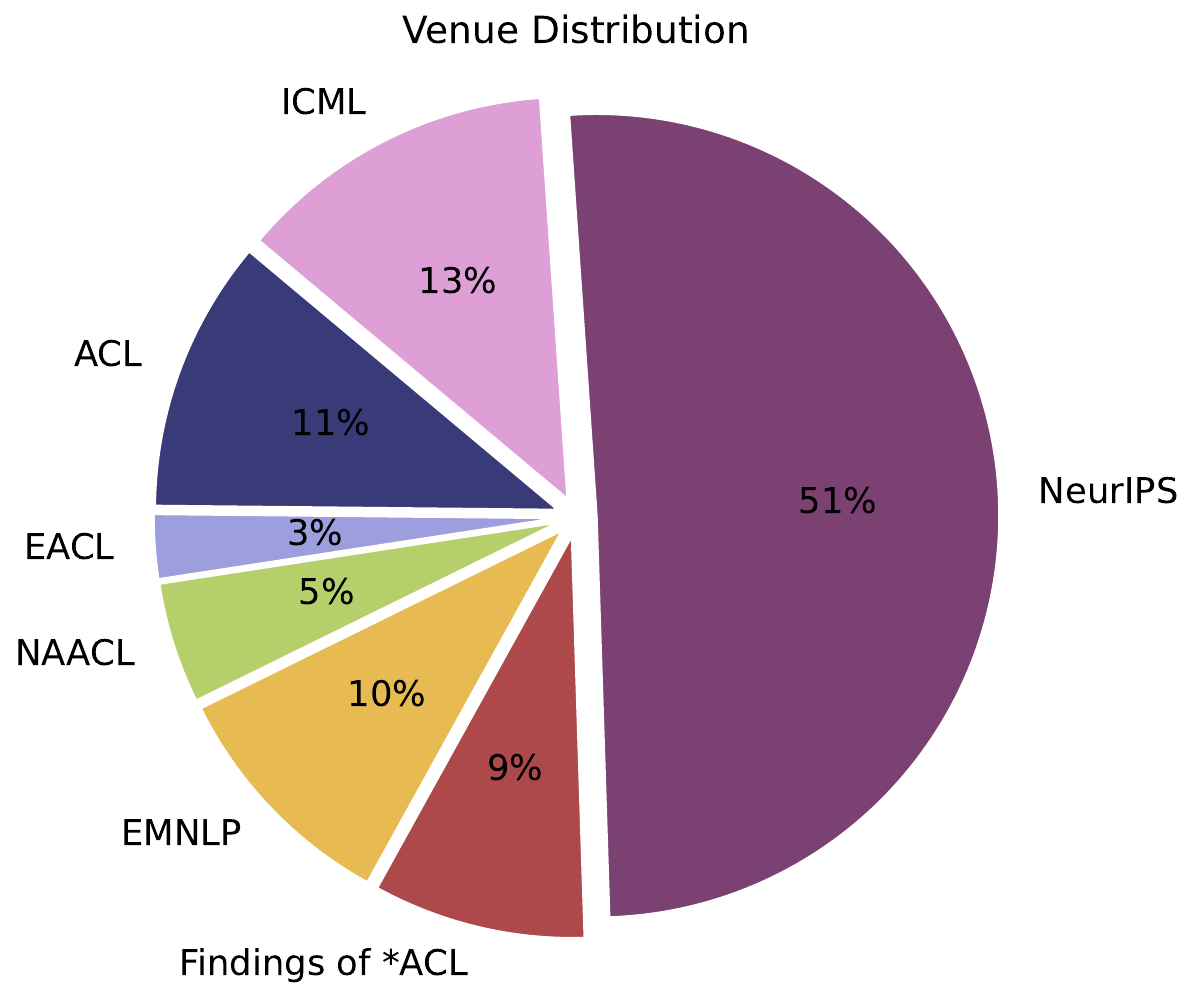}
  \caption{Venue distribution of the VISTA dataset.}
  \label{fig:venue_distribution}
\end{figure}

\paragraph{Data Splits}
After quality control, our dataset comprises 18,599 samples, with venue distributions shown in \autoref{fig:venue_distribution}. To ensure balanced domain coverage in each subset, we proportionally sample to split the dataset into training (80\%), validation (10\%), and test (10\%) sets. All subsequent experiments are conducted using these splits.

\begin{table*}[ht]
\centering
\scalebox{0.78}{
\tabcolsep=10pt
\begin{tabular}{l c c r c c c}
\toprule
\multicolumn{1}{c}{Dataset} & Language & Domain & \#Videos & \multicolumn{1}{c}{VideoLen} & SumLen\\
\midrule
MSS \citep{li-etal-2017-multi} & English, Chinese & News & 50 & 3.4 & --- \\
YouCook2 \citep{zhou2018towards} & English & Cooking & 2.0K & 5.3 & 67.8 \\
VideoStorytelling \citep{li2019video} & English & Open & 105 & 12.6 & 162.6 \\
VMSMO \citep{li-etal-2020-vmsmo} & Chinese & Social Media & 184.9K & 1.0 & 11.2 \\
MM-AVS \citep{fu-etal-2021-mm} & English & News & 2.2K & 1.8 & 56.8 \\
MLASK \citep{krubinski-pecina-2023-mlask} & Czech & News & 41.2K & 1.4 & 33.4 \\
VideoXum \citep{lin2023videoxum} & English & Activities & 14.0K & 2.1 & 49.9 \\
Shot2Story20K \citep{han2023shot2story20k} & English & Open & 20.0K & 0.3 & 201.8 \\
BLiSS \citep{he2023align} & English & Livestream & 13.3K & 5.0 & 49.0 \\
SummScreen$^{3D}$ \citep{papalampidi-lapata-2023-hierarchical3d} & English & Open & 4.5K & 40.0 & 290.0 \\
Ego4D-HCap \citep{islam2024video} & English & Open & 8.3K & 28.5 & 25.6 \\
Instruct-V2Xum \citep{hua2024v2xum} & English & Open & 30.0K & 3.1 & 239.0 \\
MMSum \citep{qiu2024mmsum} & English & Open & 5.1K & 14.5 & 21.7 \\
LfVS-T \citep{argaw2024scaling} & English & YouTube & 1.2K & 12.2 & --- \\
\rowcolor{gray!20}
VISTA (ours) &  English & Academic & 18.6K & 6.8 & 192.6 \\
\bottomrule
\end{tabular}}
\caption{Comprison of video-to-text summarization datasets. \#Videos $=$ the number of videos, whereas VideoLen and SumLen refer to the average of video duration (in minutes) and the average number of summary tokens.}
\label{tab:datasets_comparison}
\end{table*} 

\begin{figure*}[htb!]
  \centering \includegraphics[width=1\textwidth]{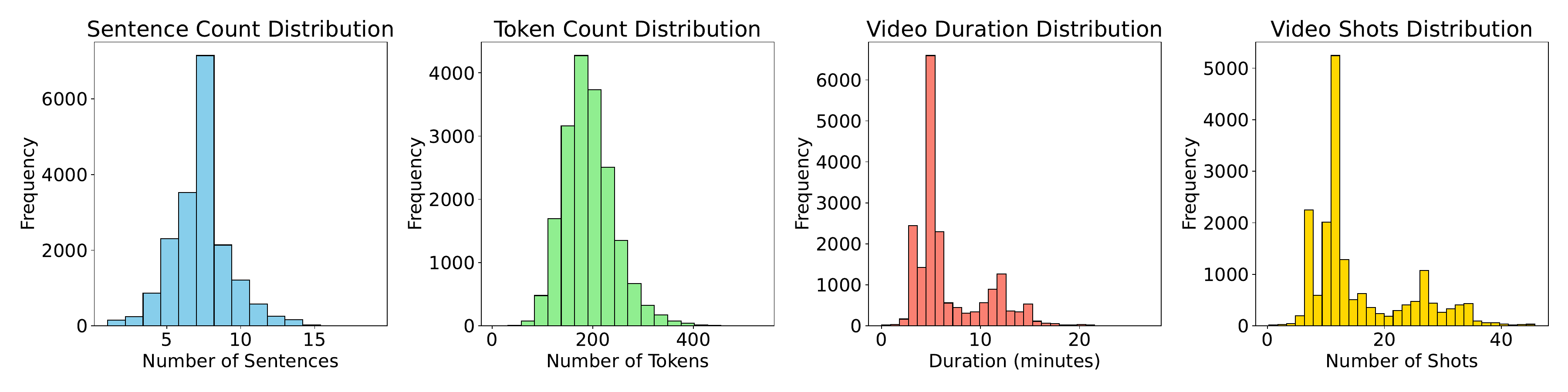}
  \caption{Distribution of summary sentences, summary tokens, video durations, and video shots in VISTA.}
  \label{fig:dataset_attribute_distributions}
\end{figure*}

\paragraph{Dataset Comparison and Statistics}

\autoref{tab:datasets_comparison} compares VISTA with several existing video-to-text summarization datasets. While many focus on open-domain (e.g., MMSum, Instruct-V2Xum) or areas like news (MLASK, MM-AVS) and activities (VideoXum), VISTA is tailored for summarizing scientific presentations. On average, it features longer inputs (6.8 minutes) than VideoXum (2.1 minutes) and MSS (3.4 minutes), as well as longer summaries (192.6 tokens), compared to YouCook2 (67.8 tokens) and VideoXum (49.9 tokens).

\begin{table}[ht]
\centering
\scalebox{0.82}{
\tabcolsep=3pt
\begin{tabular}{l r}
\toprule
Training / Validation / Test Set & 14,881 / 1,859 / 1,859 \\
\midrule
Avg. Video Length (mins) / Shots & 6.76 / 16.36 \\
\midrule
Avg. \#Summary Sent / Tokens & 7.19 / 192.62 \\
Avg. Depth of Dep Tree & 6.02 \\ 
Type-Token Ratio & 0.62 \\
Distinct-1 / -2 / -3 & 0.62 / 0.93  /  0.97 \\
\bottomrule
\end{tabular}}
\caption{Key statistics of the VISTA dataset, showcasing the average video length and shot count, summary characteristics (sentence and token counts), syntactic complexity (dependency tree depth), and lexical diversity (Type-Token Ratio and Distinct n-gram scores).}
\label{tab:statistics}
\end{table}

\autoref{tab:statistics} summarizes the VISTA dataset statistics: Videos average 6.76~minutes and 16.36~shots (we use \href{https://www.scenedetect.com/}{PySceneDetect} with \texttt{ContentDetector} to calculate video shots), while summaries contain 192.62~tokens on average across 7.19~sentences. The average dependency tree depth (Avg. Depth of Dep Tree) is 6.02, indicating the syntactic complexity of the summaries. Meanwhile, the Type-Token Ratio (TTR) is 0.62, reflecting lexical diversity. Both metrics are calculated using \href{https://spacy.io/}{spaCy}. Diversity metrics \cite{li-etal-2016-diversity}, which measure the variety of unique n-grams, yield Distinct-1, Distinct-2, and Distinct-3 scores of 0.62, 0.93, and 0.97, respectively. \autoref{fig:dataset_attribute_distributions} visualizes key attributes: Most summaries remain under 250 tokens and 10 sentences, and most videos last fewer than 10 minutes with under 30 shots. In \autoref{data_sample}, we present a random sample from the VISTA dataset.

\begin{figure*}[t]
  \centering \includegraphics[width=1\textwidth]{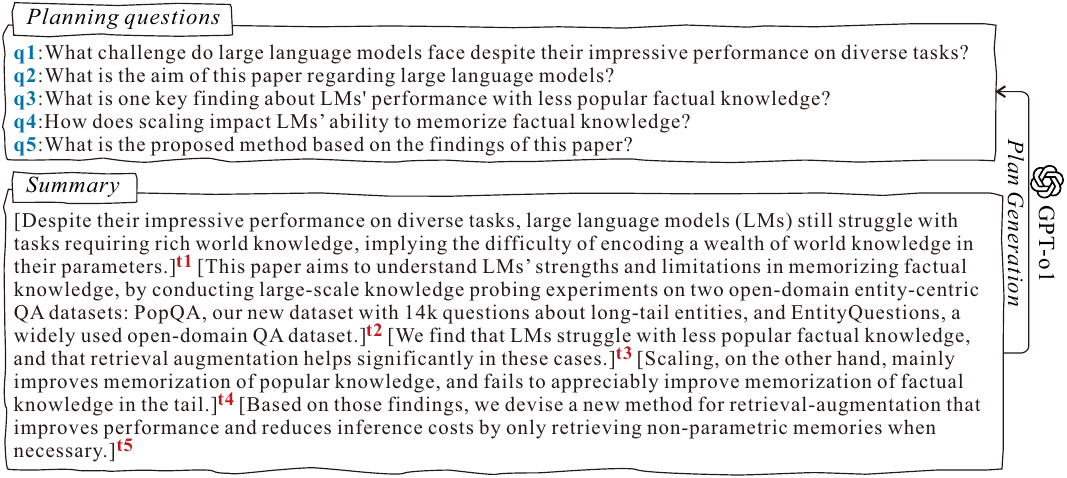}
  \caption{\texttt{GPT-o1} generates plans based on reference summaries. Each question \( q_i \) corresponds to a summary sentence  \( t_i \), which we assume constitutes its answer. Index \( i \) ranges from \( 1 \) to the number of summary sentences.}
  \label{fig: plan_extraction}
\end{figure*}

\section{Benchmarking VISTA}

\paragraph{Task Overview}

We formalize the task of summarizing recorded scientific videos as follows: Let~$v$ and~$s$ denote a video (or its transcript/audio) and its paired summary from dataset \mbox{\(D = \{(v_1, s_1), (v_2, s_2), \ldots, (v_n, s_n)\}\)}, where \( n \) signifies the number of video-summary pairs. The objective is to train a (multimodal) model \(\mathcal{M}\) to learn the conditional probability distribution \(P(s~|~v)\). Given a new video, the trained model \(\mathcal{M}\) is expected to generate an appropriate summary.

A challenge in video-to-text summarization is structuring the generated summaries in a coherent and faithful manner. Directly learning the mapping from \( v \) to \( s \) could lead to inadequate outputs, as the model lacks explicit guidance on how to organize and present the extracted information \cite{mahon-lapata-2024-modular}. Scientific abstracts often follow a relatively well-defined structure, making them suitable for a more structured generation approach \cite{takeshita-etal-2024-aclsum}. We follow previous work \cite{narayan-etal-2021-planning, narayan-etal-2023-conditional} in adopting a plan-based framework that introduces an intermediate representation to capture latent structure more effectively than simpler end-to-end approaches. Specifically, given input~$v$, we first generate a plan \( p \), which consists of a sequence of automatically generated questions \( \{q_1, q_2, \ldots, q_m\} \), each corresponding to a sentence to be verbalized in the summary. The plan explicitly controls the structure of the summary as a whole and the content of each of its sentences (which are meant to answer the questions in the plan). The model is then trained to learn the extended conditional probability distribution \( P(s~|~v,p) \), ensuring that the generated summaries follow the structure and flow of plan~$p$.

\paragraph{Plan Generation}
\label{sec: plan_generation}
We hypothesize that summary sentences can be viewed as responses to plan questions, where the plan consists of an ordered sequence of questions directly associated with the target content. This idea is inspired by the theory of Question Under Discussion (QUD; \citet{roberts2012information, wu-etal-2023-elaborative, suvarna-etal-2024-qudselect}), which posits that discourse often revolves around a set of questions that guide the structure and interpretation of the conversation.

We leverage \texttt{GPT-o1} \cite{achiam2023gpt} to generate silver-standard plans based on reference summary sentences and their preceding context. As shown in \autoref{fig: plan_extraction}, for example, question \(q_3\) is generated based on target sentence~\(t_3\) and the summary sentences preceding it (i.e.,~\(t_1\) and \(t_2\)), and so on. As a result, the question sequence preserves the order of sentences in the reference summaries, ensuring that the plan maintains a natural and coherent flow consistent with the structure of reference summaries. The prompt used to generate plan questions is provided in Appendix \autoref{plan_question_generation}. We discuss the quality of the silver-standard plans through manual investigation in \autoref{plan_quality_validation}.

\paragraph{Summarization Model}
We train two independent modules corresponding to Plan Generation (\texttt{PG}) and  Summary Generation (\texttt{SG}). The \texttt{PG} module is trained on pairs of  \( (v, p) \)~samples. The \texttt{SG} module is trained on tuples \( ([v; p], s) \), where \( [v; p] \) is the concatenation of the input~\( v \) and its plan~\( p \). During inference, the trained \texttt{PG} module predicts plan \( \hat{p} \) for input~\( v \), and the tuple \( [v; \hat{p}] \) is fed into the \texttt{SG} module to generate the final summary. Both modules have the same backbone but are trained independently.

\begin{table*}[t]
\centering
\scalebox{0.65}{
\tabcolsep=5pt
\begin{threeparttable}
\begin{tabular}{c l c c c c c c c c c c}
\toprule
Method & Model & Open-source & R1 & R2 & RLsum & SacreBLEU & Meteor & BERTscore & CIDEr-D & VideoScore & FactVC\\
\hline
\multirow{12}*{\rotatebox[origin=c]{90}{Zero-shot Learning}}
& \texttt{LLaMA-3.1$_{transcript}$} & \cmark & 23.68 & 4.22 & 21.39 & 2.70 & 14.62 & 80.93 & 1.17 & 1.53 & 34.32 \\
~ & \texttt{LLaMA-3.1$_{OCR}$} & \cmark & 24.02 & 4.37 & 21.42 & 2.63 & 14.59 & 80.33 & 1.19 & 1.50 & 34.06 \\
~ & \texttt{Qwen2-Audio} & \cmark & 23.52 & 4.29 & 21.53 & 2.49 & 14.77 & 80.62 & 1.15 & 1.59 & 34.31 \\
\cdashline{2-11}
~ & \texttt{Claude 3.5 Sonnet} & \xmark & 27.71 & 5.59 & 24.14 & 3.14 & 17.53 & 82.57 & 1.32 & 1.91 & 50.11 \\
~ & \texttt{Gemini 2.0} & \xmark & 27.82 & 5.66 & 24.29 & 4.22 & 17.83 & 82.64 & 1.47 & 2.02 & 52.02 \\
~ & \texttt{GPT-o1} & \xmark & 27.90 & 5.69 & 24.37 & 4.38 & 17.90 & 82.63 & 1.61 & 2.17 & 51.36 \\
~ & \texttt{Video-LLaMA} & \cmark & 20.18 & 3.19 & 21.24 & 1.76 & 13.73 & 81.31 & 1.08 & 1.63 & 32.25 \\
~ & \texttt{Video-ChatGPT}  & \cmark & 20.36 & 3.52 & 21.43 & 1.79 & 14.01 & 81.35 & 1.11 & 1.63 & 33.21 \\
~ & \texttt{Video-LLaVA}  & \cmark & 25.29 & 4.50 & 22.52 & 2.82 & 15.13 & 81.39 & 1.17 & 1.65 & 36.45 \\
~ & \texttt{LLaMA-VID}  & \cmark & 25.31 & 4.77 & 22.53 & 2.88 & 15.27 & 81.32 & 1.14 & 1.64 & 36.39 \\
~ & \texttt{LLaVA-NeXT-Interleave}  & \cmark & 25.41 & 4.82 & 22.68 & 2.92 & 15.25 & 81.40 & 1.18 & 1.73 & 40.12 \\
~ & \texttt{mPLUG-Owl3} & \cmark & 25.57 & 4.82 & 22.84 & 2.99 & 15.33 & 81.39 & 1.21 & 1.77 & 42.07 \\
~ & \texttt{\mbox{Plan-mPlug-Owl3}}$^\divideontimes$ & \cmark & \cellcolor{green} \textbf{25.62}$^\dag$ & \cellcolor{green} \textbf{4.95}$^\dag$$^\ddag$ & \cellcolor{green} \textbf{22.97}$^\dag$$^\ddag$ & \cellcolor{green} \textbf{3.14}$^\dag$$^\ddag$ & \cellcolor{green} \textbf{15.39}$^\dag$$^\ddag$ & \cellcolor{green} \textbf{81.45}$^\ddag$ & \cellcolor{green} \textbf{1.27}$^\dag$$^\ddag$ & \cellcolor{green} \textbf{1.86}$^\dag$$^\ddag$ & \cellcolor{green} \textbf{47.37}$^\dag$$^\ddag$\\
\midrule
\multirow{10}*{\rotatebox[origin=c]{90}{QLoRA Fine-tuning}}
& \texttt{LLaMA-3.1$_{transcript}$} & \cmark & 32.24 & 11.38 & 30.39 & 8.03 & 21.57 & 82.39 & 3.86 & 2.81 & 53.22 \\
~ & \texttt{LLaMA-3.1$_{OCR}$} & \cmark & 33.01 & 12.11 & 30.52 & 8.04 & 21.55 & 82.41 & 3.92 & 2.77 & 53.19 \\
~ & \texttt{Qwen2-Audio} & \cmark & 32.17 & 12.05 & 30.77 & 7.87 & 21.86 & 82.36 & 4.11 & 2.80 & 54.27 \\
\cdashline{2-11}
~ & \texttt{Video-LLaMA} & \cmark & 30.74 & 9.44 & 28.33 & 6.45 & 22.49 & 82.61 & 3.99 & 2.77 & 52.05 \\
~ & \texttt{Video-ChatGPT}  & \cmark & 31.68 & 10.50 & 30.40 & 7.63 & 23.67 & 82.62 & 4.02 & 2.78 & 55.02 \\
~ & \texttt{Video-LLaVA}  & \cmark & 33.16 & 12.64 & 30.37 & 8.17 & 23.92 & 82.81 & 4.26 & 2.83 & 59.13 \\
~ & \texttt{LLaMA-VID}  & \cmark & 33.31 & 12.73 & 30.49 & 8.22 & 23.90 & 83.01 & 4.31 & 2.88 & 62.20 \\
~ & \texttt{LLaVA-NeXT-Interleave}  & \cmark & 33.37 & 12.77 & 30.56 & 8.30 & 23.95 & 83.47 & 4.47 & 2.93 & 66.14 \\
~ & \texttt{mPLUG-Owl3} & \cmark & 33.40 & 12.82 & 30.66 & 8.29 & 23.97 & 83.49 & 4.47 & 2.92 & 70.08 \\
~ & \texttt{\mbox{Plan-mPlug-Owl3}} & \cmark & \cellcolor{green} \textbf{33.52}$^\dag$$^\ddag$ & \cellcolor{green} \textbf{13.01}$^\dag$$^\ddag$ & \cellcolor{green} \textbf{31.10}$^\dag$$^\ddag$ & \cellcolor{green} \textbf{8.33} & \cellcolor{green} \textbf{24.11}$^\dag$$^\ddag$ & \cellcolor{green} \textbf{83.53}$^\dag$ & \cellcolor{green} \textbf{4.52} & \cellcolor{green} \textbf{3.11}$^\dag$$^\ddag$ & \cellcolor{green} \textbf{73.11}$^\dag$$^\ddag$ \\
\midrule
\multirow{10}*{\rotatebox[origin=c]{90}{Full Fine-tuning}}
& \texttt{LLaMA-3.1$_{transcript}$} & \cmark & 33.37 & 11.93 & 30.86 & 8.27 & 25.12 & 83.71 & 4.87 & 3.21 & 63.38 \\
~ & \texttt{LLaMA-3.1$_{OCR}$} & \cmark & 34.02 & 12.42 & 31.72 & 8.51 & 15.11 & 84.09 & 4.89 & 3.32 & 65.84 \\
~ & \texttt{Qwen2-Audio} & \cmark & 33.82 & 12.37 & 31.63 & 8.33 & 25.09 & 83.62 & 4.83 & 3.22 & 66.62 \\
\cdashline{2-11}
~ & \texttt{Video-LLaMA} & \cmark & 32.19 & 11.86 & 31.68 & 8.41 & 24.99 & 83.83 & 4.77 & 3.04 & 64.21 \\
~ & \texttt{Video-ChatGPT}  & \cmark & 32.47 & 12.11 & 32.21 & 8.72 & 25.09 & 83.91 & 4.82 & 3.11 & 66.09 \\
~ & \texttt{Video-LLaVA}  & \cmark & 33.28 & 13.39 & 32.78 & 9.10 & 25.42 & 83.97 & 4.87 & 3.13 & 66.12 \\
~ & \texttt{LLaMA-VID}  & \cmark & 33.47 & 13.53 & 32.80 & 9.21 & 25.41 & 84.03 & 4.91 & 3.17 & 68.30 \\
~ & \texttt{LLaVA-NeXT-Interleave}  & \cmark & 33.75 & 13.61 & 32.88 & 9.26 & 25.63 & 84.11 & 5.01 & 3.23 & 73.42 \\
~ & \texttt{mPLUG-Owl3} & \cmark & 34.22 & 13.62 & 32.91 & 9.32 & 25.72 & 84.22 & 5.03 & 3.28 & 71.94 \\
~ & \texttt{\mbox{Plan-mPlug-Owl3}} & \cmark & \cellcolor{green} \textbf{34.53}$^\dag$$^\ddag$ & \cellcolor{green} \textbf{13.74}$^\dag$$^\ddag$ & \cellcolor{green} \textbf{33.25}$^\dag$$^\ddag$ & \cellcolor{green} \textbf{9.56}$^\dag$$^\ddag$ & \cellcolor{green} \textbf{25.88}$^\dag$$^\ddag$ & \cellcolor{green} \textbf{84.37}$^\dag$$^\ddag$ & \cellcolor{green} \textbf{5.15}$^\dag$$^\ddag$ & \cellcolor{green} \textbf{3.33}$^\dag$$^\ddag$ & \cellcolor{green} \textbf{75.41}$^\dag$$^\ddag$ \\
\bottomrule
\end{tabular}
\end{threeparttable}
}
\caption{Model performance on VISTA dataset. In \texttt{\mbox{Plan-mPlug-Owl3}}$^\divideontimes$, only the \textsc{PG} module is trained. Plans generated by the \textsc{PG} on the test set serve as input to the \textsc{SG} module for zero-shot inference (no training is applied to the \textsc{SG} module).  Symbols~$^\dag$ and $^\ddag$ indicate that the performance of \PlanSum is significantly ($p<0.05$) different from  \texttt{LLaVA-NeXT-Interleave} (third best) and \texttt{mPLUG-Owl3} (second best), when using the paired t-test.}
\label{tab:model_performance}
\end{table*}

\section{Experiments}

\paragraph{Baseline Models} We benchmark our dataset using three learning settings: Zero-shot learning, QLoRA fine-tuning \cite{dettmers2024qlora}, and full-parameter fine-tuning. For zero-shot learning, we test closed-source multimodal models, including \texttt{GPT-o1} \cite{achiam2023gpt}, \texttt{Gemini 2.0} \cite{team2023gemini}, \texttt{ \texttt{Claude 3.5 Sonnet} } \cite{anthropic_claude3_5_sonnet}, as well as open-source video LMMs such as \texttt{Video-LLaMA} \cite{zhang-etal-2023-video}, \texttt{Video-ChatGPT} \cite{maaz-etal-2024-video}, \texttt{Video-LLaVA} \cite{lin-etal-2024-video}, \texttt{LLaMA-VID} \cite{li2025llama}, \texttt{LLaVA-NeXT-Interleave} \cite{li2024llava}, and \texttt{mPLUG-Owl3} \cite{ye2024mplug}. These open-source video LMMs process videos by extracting multimodal features, such as visual and/or audio components, using cross-modal attention mechanisms to align and integrate information across modalities. 

We also assess \texttt{LLaMA-3.1} \cite{touvron2023llama} and \texttt{Qwen2-Audio} \cite{chu2024qwen2} to examine if text- or audio-based models can accomplish the summarization task without taking video information into account. For \texttt{LLaMA-3.1}, we explore two variants: In \texttt{LLaMA-3.1$_{transcript}$}, we extract audio from video files using \href{https://zulko.github.io/moviepy/}{\texttt{moviepy}} and transcribe it with OpenAI's \href{https://github.com/openai/whisper}{\texttt{Whisper-1}} to generate text input for the model. In \texttt{LLaMA-3.1$_{OCR}$}, we apply \href{https://github.com/JaidedAI/EasyOCR}{\texttt{EasyOCR}} to extract on-screen text from video frames and use the OCR-generated text as input for summarization. Similarly, for \texttt{Qwen2-Audio}, we use \href{https://zulko.github.io/moviepy/}{moviepy} to convert video files into audio and treat the audio as input. Exact model versions are provided in \autoref{model_versions}. Based on our benchmarking results, we select the best-performing model as the backbone for the plan-based strategy and evaluate its performance. Prompts for the above models are offered in \autoref{sec:prompts-study} (Figures~\ref{non_plan_based_summary_generation}--\ref{SG_model}).

\paragraph{Experimental Setup}
To ensure a fair comparison, all models, including baselines, plan-based models, and ablation models, are evaluated under identical hyperparameter settings unless explicitly stated otherwise. All models are tested using identical prompt instructions. Detailed hyperparameter configurations are presented in \autoref{hyper-parameters_settings}.

\paragraph{Evaluation Metrics}
We report a set of evaluation metrics to measure informativeness, alignment, and factual consistency in summaries. For informativeness, we utilize ROUGE \cite{lin-2004-rouge}, SacreBLEU \cite{post-2018-call}, METEOR \cite{banerjee-lavie-2005-meteor}, BERTScore \cite{zhang2019bertscore}, and CIDEr-D \cite{vedantam2015cider}. Specifically, we provide the F1 scores for Rouge-1 (R1), Rouge-2 (R2), and Rouge-LSum (RLSUM). Alignment to the input video is evaluated with VideoScore \cite{he-etal-2024-videoscore}, and factual consistency with FactVC \cite{liu-wan-2023-models}. Detailed descriptions of these metrics are given in \autoref{automatic_evaluation_metrics}.

\section{Results and Analysis}
\label{results_analysis}

\paragraph{General Results} \autoref{tab:model_performance} compares model performance across three learning settings: Zero-shot, QLoRA fine-tuning, and full-parameter fine-tuning. Overall, fine-tuning on in-domain data yields substantial performance gains across all evaluation metrics. Full fine-tuning consistently outperforms QLoRA. While closed-source models such as \texttt{GPT-o1} and \texttt{Gemini} typically lead in zero-shot performance, open-source models like \texttt{mPLUG-Owl3} and \PlanSum achieve competitive or even superior results when fine-tuned, especially in semantic alignment (BERTScore) and video-text consistency (VideoScore).

We observe that video-based LMMs consistently outperform text-based and audio-based models. While models such as \texttt{LLaMA-3.1$_{transcript}$}, \texttt{LLaMA-3.1$_{OCR}$}, and \texttt{Qwen2-Audio} yield comparable results, they lag behind video-grounded models in overall performance. In particular, \texttt{mPLUG-Owl3} achieves SOTA results across most metrics, highlighting the crucial role of visual information in enhancing summarization quality.

\PlanSum is the plan-based approach built on \texttt{mPLUG-Owl3}, outperforming all open-source baselines in both zero-shot and fine-tuned settings. For zero-shot inference, the \texttt{\mbox{Plan-mPlug-Owl3}}$^\divideontimes$ variant, which fine-tunes only the Plan Generation (PG) module, surpasses other models in summary quality, factual consistency, and semantic alignment. With full-parameter fine-tuning, \PlanSum achieves the highest overall scores across models, showing improvements in factual accuracy (+3.47 in FactVC) and quality (+0.34 in RLsum) compared to \texttt{mPLUG-Owl3}. However, all models (including the plan-based method) exhibit hallucinations (FactVC) and alignment (VideoScore) issues, and there are still significant differences (p-value of the paired t-test is less than 0.05) between the human performance in this task, with reference summaries scoring 88.54 on FactVC and 4.62 on VideoScore.

\paragraph{Impact of Modality Interplay} 

To explore the impact of different modality combinations on our multimodal tasks, we conduct an experiment using Video-LLaMA \cite{zhang-etal-2023-video}. Seven modality combinations are considered, including unimodal inputs (video, audio, transcript) and their pairwise or joint combinations. For each configuration, only the corresponding modality modules are updated while the remaining ones are kept frozen. The summarized results are shown in \autoref{tab:modality-analysis}.

\begin{table*}[h]
\centering
\resizebox{\textwidth}{!}{
\begin{tabular}{l|cccc|cccc|cccc}
\toprule
\multirow{2}{*}{Modality} & \multicolumn{4}{c|}{Zero-shot Learning} & \multicolumn{4}{c|}{QLoRA Fine-tuning} & \multicolumn{4}{c}{Full Fine-tuning} \\
 & R2 & RLsum & VideoScore & FactVC & R2 & RLsum & VideoScore & FactVC & R2 & RLsum & VideoScore & FactVC \\
\midrule
Video only & 2.68 & 20.34 & 1.55 & 28.93 & 8.83 & 27.51 & 2.65 & 50.66 & 10.78 & 30.02 & 2.91 & 60.87 \\
Audio only & 2.14 & 19.72 & 1.41 & 26.84 & 7.52 & 26.34 & 2.48 & 45.79 & 9.23 & 27.93 & 2.73 & 58.02 \\
Transcript only & 2.02 & 18.01 & 1.34 & 25.53 & 6.91 & 24.33 & 2.39 & 44.87 & 8.44 & 25.81 & 2.35 & 54.11 \\
Video + Audio & \textbf{3.19} & \textbf{21.24} & \textbf{1.63} & \textbf{32.25} & \textbf{9.44} & \textbf{28.33} & \textbf{2.77} & \textbf{52.05} & \textbf{11.86} & \textbf{31.68} & \textbf{3.04} & \textbf{64.21} \\
Video + Transcript & 1.87 & 18.94 & 1.39 & 27.76 & 7.35 & 24.82 & 2.51 & 48.63 & 9.01 & 27.19 & 2.65 & 58.91 \\
Audio + Transcript & 1.64 & 18.55 & 1.35 & 27.48 & 7.23 & 24.73 & 2.38 & 47.15 & 8.57 & 25.82 & 2.54 & 55.39 \\
Video + Audio + Transcript & 1.92 & 19.13 & 1.47 & 28.60 & 7.37 & 25.29 & 2.52 & 50.72 & 9.22 & 27.21 & 2.61 & 59.30 \\
\bottomrule
\end{tabular}
}
\caption{Performance comparison of different modality combinations.}
\label{tab:modality-analysis}
\end{table*}

The results consistently show that video is the strongest standalone modality, likely due to its rich spatial-temporal information. Audio offers complementary prosodic and timing cues, but lacks semantic visual grounding. The transcript, while semantically rich, often introduces long, noisy, and unstructured inputs, particularly from ASR systems, that can overwhelm the model’s attention and interfere with alignment. These findings suggest that current video-based LMMs face challenges in effectively aligning and fusing token-heavy, noisy textual inputs with corresponding visual or audio information.

\paragraph{Impact of Plan Generation Ablations}
\label{plan_generation_stratege}
We analyze the plan generation ablation by comparing it with simpler baselines: Lead-3$_Q$, Tail-3$_Q$, and Random-3$_Q$. In these ablation baselines, plans are generated by selecting the first three, last three, or three randomly chosen summary sentences, respectively. Each selected sentence serves as a target for generating a question, with its preceding sentences providing the context. For instance, in the Lead-3$_Q$ setting, the first sentence is used as the target (without any preceding context), prompting the first question in the plan, while subsequent sentences incorporate earlier ones as context. Additionally, we compare the case where QUD is not considered. That is, we directly let \texttt{GPT-o1} generate all plan questions at once based on the reference summary (\texttt{NoQUD}).

\begin{table}[t]
\centering
\scalebox{0.82}{
\tabcolsep=4pt
\begin{threeparttable}
\begin{tabular}{l c c c c}
\toprule
Model & R2 & RLsum & VideoScore & FactVC\\
\hline
\PlanSum & 13.74 & 33.25 & 3.33 & 75.41 \\
\hdashline
\texttt{NoQUD} & 13.66 & 33.02 & 3.28 & 73.32 \\
% \hdashline
Lead-3$_Q$ & 12.87 & 30.64 & 2.95 & 71.26\\
Tail-3$_Q$ & 11.62 & 30.51 & 2.88 & 63.82\\
Random-3$_Q$ & 11.57 & 30.48 & 2.87 & 64.28\\
\bottomrule
\end{tabular}
\end{threeparttable}
}
\caption{Performance comparison of different plan generation ablations under full fine-tuning settings.}
\label{tab:ablation_study}
\end{table}

\autoref{tab:ablation_study} underlines the performance differences across different plan generation ablations. For \texttt{NoQUD}, it underperforms compared to the QUD-based approach. The Lead-3$_Q$ strategy performs better overall compared to Tail-3$_Q$ and Random-3$_Q$, indicating that initial sentences offer stronger contextual continuity for generating plan questions.

\paragraph{Impact of Plan Quality}

We assess how the quality of the plan questions affects model performance. We apply \texttt{GPT-o1} as a question generator in a zero-shot setting in our previous experiments. For comparative analysis, we additionally incorporate \texttt{Llama-3.1} and a state-of-the-art question generation algorithm (RAST) from \citet{gou-etal-2023-diversify} to generate the plan questions. In addition, we apply a Random Replacement (RR) method, where questions generated by \texttt{GPT-o1} are randomly replaced with irrelevant ones. The number of replaced questions per summary ranges from one to the entire set. We also introduce full random replacement (FRR), where questions generated by \texttt{GPT-o1} are all replaced with random irrelevant questions.\footnote{The prompt for generating irrelevant questions is given in Appendix~\autoref{Irrelevant_Question_Generation}.}

\begin{figure}[t]
  \centering 
  \includegraphics[width=0.48\textwidth]{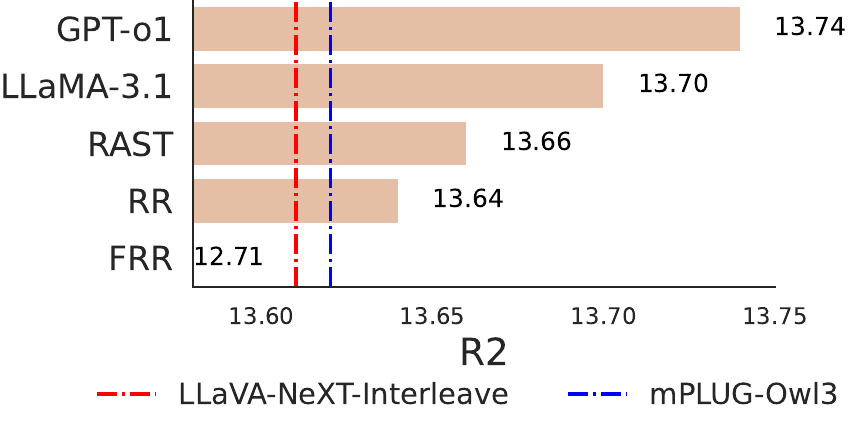}
  \caption{Noise in plan generation impacts summarization performance. FRR is a shorthand for Full Random Replacement, and RR for Random Replacement. RAST is a SOTA question generation method.}
  \label{fig:plan_quality_impact}
\end{figure}

\autoref{fig:plan_quality_impact} reveals that the quality of plan questions does influence the summarization performance: Using \texttt{GPT-o1} to generate questions outperforms the rest. The FRR method performs the worst, as irrelevant questions disrupt the alignment between the plan and summary content. We also find that the plan-based method exhibits a certain degree of robustness, as it performs reasonably well even when the plans contain some degree of noise (RR vs. FRR). These findings emphasize the importance of question relevance and quality in structuring the output summaries.

\paragraph{Planning Beyond Vision}
\label{generalizability_non_visual}

While our primary objective is to evaluate the planning framework in the context of video-to-text summarization, it is valuable to assess its applicability to unimodal, non-visual models. To this end, we conduct supplementary experiments applying the planning method to three models that do not utilize video inputs: (1) \texttt{LLaMA-3.1$_{transcript}$} (ASR-based textual input), (2) \texttt{LLaMA-3.1$_{OCR}$} (OCR-based textual input), and (3) \texttt{Qwen2-Audio} (audio-based input). For each model, we compare baseline performance (i.e., without planning) against the planning counterpart. As summarized in Table~\ref{tab:planning-nonvideo}, planning consistently improves performance across all settings and evaluation metrics. A paired t-test confirms that these improvements are statistically significant ($p < 0.05$).

These findings demonstrate that the planning method does not function solely as a domain-specific enhancement but rather as a generalizable scaffold that supports better discourse structure, even in the absence of visual input. We hypothesize that, for text- and audio-based models, planning mitigates the lack of spatial-temporal signals by providing discourse-level anchors, such as intent-driven prompts (e.g., ``What problem is being addressed?''), that guide the model’s summarization trajectory.

\begin{table*}[ht]
\centering
\resizebox{\textwidth}{!}{
\begin{tabular}{c|c|c|c|c|c}
\toprule
Model & Setting & R2 & RLsum & VideoScore & FactVC \\
\midrule
\multirow{3}{*}{\texttt{LLaMA-3.1$_{transcript}$}} 
& Zero-shot Learning & 4.22 $\rightarrow$ \textbf{4.56} & 21.39 $\rightarrow$ \textbf{22.01} & 1.53 $\rightarrow$ \textbf{1.75} & 34.32 $\rightarrow$ \textbf{40.78} \\
& QLoRA Fine-tuning & 11.38 $\rightarrow$ \textbf{11.62} & 30.39 $\rightarrow$ \textbf{30.55} & 2.81 $\rightarrow$ \textbf{3.02} & 53.22 $\rightarrow$ \textbf{60.47} \\
& Full Fine-tuning & 11.93 $\rightarrow$ \textbf{12.24} & 30.86 $\rightarrow$ \textbf{31.38} & 3.21 $\rightarrow$ \textbf{3.25} & 63.38 $\rightarrow$ \textbf{65.21} \\
\midrule
\multirow{3}{*}{\texttt{LLaMA-3.1$_{OCR}$}} 
& Zero-shot Learning & 4.37 $\rightarrow$ \textbf{4.59} & 21.42 $\rightarrow$ \textbf{21.89} & 1.50 $\rightarrow$ \textbf{1.72} & 34.06 $\rightarrow$ \textbf{40.24} \\
& QLoRA Fine-tuning & 12.11 $\rightarrow$ \textbf{12.33} & 30.52 $\rightarrow$ \textbf{30.78} & 2.77 $\rightarrow$ \textbf{2.98} & 53.19 $\rightarrow$ \textbf{60.38} \\
& Full Fine-tuning & 12.42 $\rightarrow$ \textbf{12.75} & 31.72 $\rightarrow$ \textbf{32.19} & 3.32 $\rightarrow$ \textbf{3.38} & 65.84 $\rightarrow$ \textbf{67.53} \\
\midrule
\multirow{3}{*}{\texttt{Qwen2-Audio}} 
& Zero-shot Learning & 4.29 $\rightarrow$ \textbf{4.51} & 21.53 $\rightarrow$ \textbf{22.18} & 1.59 $\rightarrow$ \textbf{1.77} & 34.31 $\rightarrow$ \textbf{40.52} \\
& QLoRA Fine-tuning & 12.05 $\rightarrow$ \textbf{12.19} & 30.77 $\rightarrow$ \textbf{31.04} & 2.80 $\rightarrow$ \textbf{3.01} & 54.27 $\rightarrow$ \textbf{61.44} \\
& Full Fine-tuning & 12.37 $\rightarrow$ \textbf{12.68} & 31.63 $\rightarrow$ \textbf{32.12} & 3.22 $\rightarrow$ \textbf{3.25} & 66.62 $\rightarrow$ \textbf{68.25} \\
\bottomrule
\end{tabular}
}
\caption{Performance of baseline vs. planning models in non-video settings across different learning regimes. Each cell shows the result \textit{before $\rightarrow$ after} applying the planning method.}
\label{tab:planning-nonvideo}
\end{table*}

Notably, despite these gains, video-based planning models such as \texttt{Plan-mPLUG-Owl3} still outperform their non-visual counterparts by a notable margin. Nonetheless, our findings reinforce the idea that structured planning improves summarization quality beyond the video domain. In \autoref{video_context_impact}, we further explore the effect of video content on our summarization task, varying the length of the video given as input to the model. We also perform experiments with different textual contexts for generating plan questions in \autoref{text_context_impact}, and with controlled generation in \autoref{controllable_generation}. Additionally, we present an error analysis of model output in \autoref{case_study}.

\section{Human Evaluation}
\label{human_evaluation}
We conduct a human evaluation on 50 randomly selected instances from the VISTA test set. Annotators include master’s and doctoral students in computer science or computational linguistics with advanced English proficiency. They receive compensation per our university’s standard rate and are blind to the source of each summary to ensure impartial assessment. We compare \PlanSum, \texttt{mPLUG-Owl3}, \texttt{LLAVA-NeXT-Interleave}, and \texttt{GPT-o1} against human reference summaries. Three independent annotators are asked to review the source video and evaluate corresponding model outputs (and the human upper bound) on a 1--5 Likert scale for Faithfulness, Relevance, Informativeness, Conciseness, and Coherence (higher scores indicate better quality). They are also asked to provide an overall ranking. In total, participants rated~750 samples ($50 \times 5 \times 3$). \autoref{human_evaluation_guideline}  contains the full evaluation instructions.

\begin{figure}[t]
  \centering \includegraphics[width=0.48\textwidth]{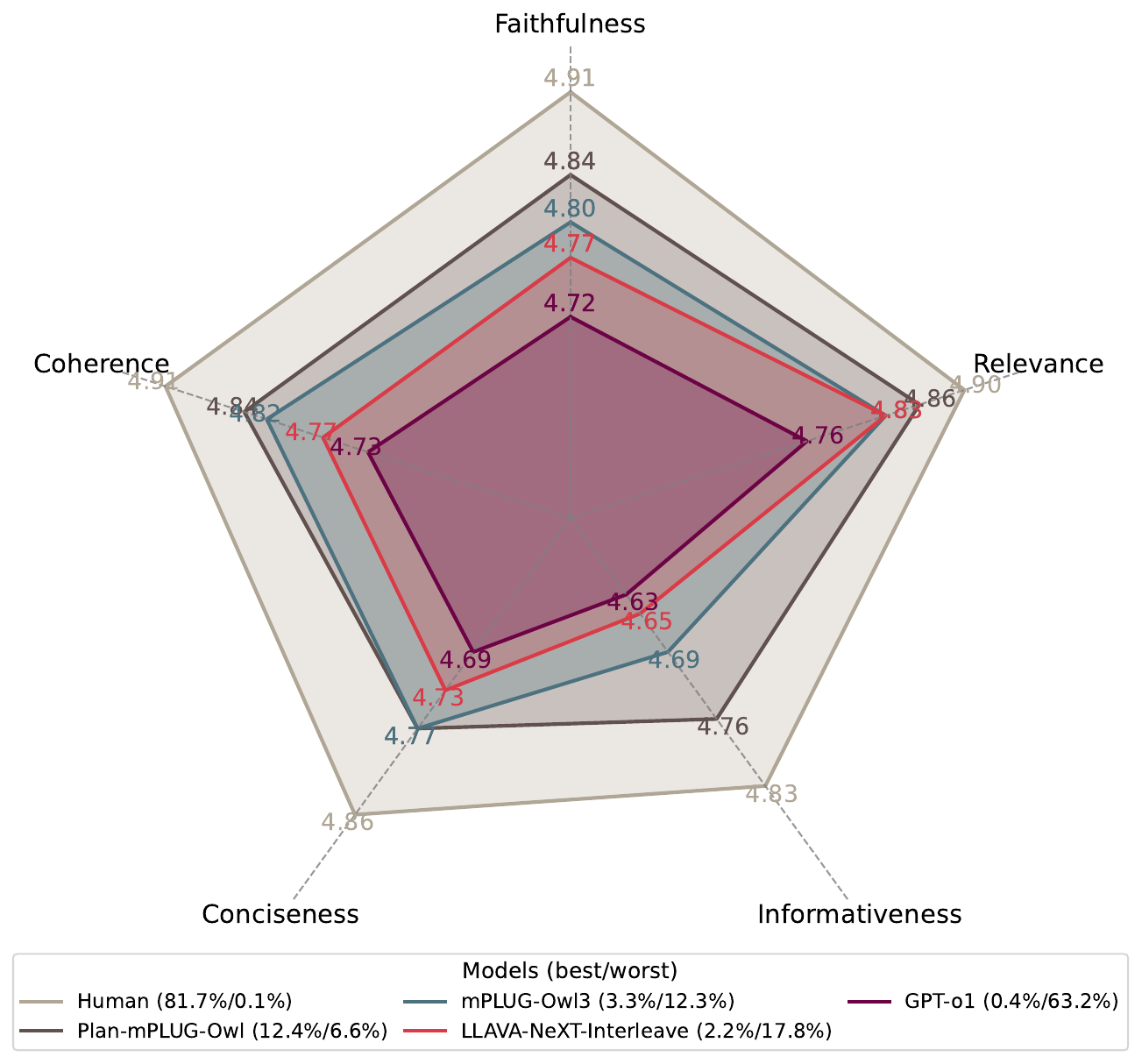}
  \caption{Human evaluation results. Human-written summaries consistently outperform all neural models.}
  \label{fig:human_evaluation}
\end{figure}

\autoref{fig:human_evaluation} presents the performance of each model, along with the proportion of instances where models are rated best or worst. Fleiss’ Kappa scores for Faithfulness (\(\kappa = 0.767\)), Relevance (\(\kappa = 0.842\)), Informativeness (\(\kappa = 0.721\)), Conciseness (\(\kappa = 0.792\)), and Coherence (\(\kappa = 0.813\)) indicate a substantial level of agreement, with an average agreement score of \(\kappa = 0.787\). Overall, human-written summaries outperform all neural summarization models in quality, as they are perceived as substantially more faithful, coherent, concise, and informative. Human-written summaries are 81.7\% more likely to be rated as best compared to model-generated summaries. 

Among the four neural models, \texttt{GPT-o1} performs worst, being rated as worst~63.2\% of the time. \texttt{LLAVA-NeXT-Interleave} follows suit, with a 17.8\%~chance of receiving the worst ranking. The plan-based model, \texttt{Plan-mPLUG-Owl3}, outperforms \texttt{mPLUG-Owl3} and demonstrates superior performance across all metrics. Additionally, it stands out among neural summarization systems for its higher likelihood of generating high-quality summaries. Paired t-tests show that human answers are considered significantly better than all neural models in all metrics (\(p < 0.05\)), revealing a clear gap between automatic systems and human performance on the VISTA dataset. The plan-based method is significantly better (\(p < 0.05\)) than other neural models in faithfulness, coherence, and informativeness, although it falls short of human performance. We also evaluate all samples of the test set with an LMM-as-Judge and obtain results that are broadly consistent with human evaluation. We describe the details of this study in \autoref{lmm-as-judge}.

\section{Conclusion}
This paper introduces VISTA, a novel dataset specifically curated for the task of summarizing scientific video presentations into concise and coherent textual summaries. Comprehensive evaluations across multiple large (language/audio/multimodal) models demonstrate that this task poses significant challenges due to the complexity and multimodal nature of scientific presentations. To address these challenges, we operate a plan-based summarization approach that incorporates discourse-aware planning prior to summary generation. This method consistently improves summary quality, factual coverage, and coherence across multiple settings. In addition to presenting the dataset, our study reveals that even the strongest current models still fall short of matching human performance by a noticeable margin. We believe that VISTA could provide a robust and extensible foundation for future research on video-to-text summarization.

\section*{Ethical Considerations}
All data in our dataset are sourced from publicly accessible resources, strictly adhering to relevant copyright regulations. Each data sample explicitly includes the corresponding source URL and author attribution. Throughout the processes of data processing, experimental analysis, model training, and evaluation, no instances of privacy infringement were identified. In human evaluations, all participants volunteered willingly and were fairly compensated. We provided a safe and comfortable environment for our participants and complied with \href{https://www.aclweb.org/adminwiki/index.php/ACL_Policy_on_Publication_Ethics}{ACL's Policy on Publication Ethics} throughout our studies.

\section*{Limitations}

\paragraph{Data} All the summary and video data used in this study are open source. While our sources are generally of high quality and exhibit a broad range of diversity, we have not investigated inherent biases in the data. Moreover, as these data represent only a small fraction of real-world data, our findings may not extend to all video-to-text summarization scenarios.

\paragraph{Task} In our task, we consider the paper abstract as a proxy for the summary of the corresponding video. This hypothesis has been supported by our two-stage quality control process, which ensures a strong alignment. However, we acknowledge that there may be nuanced differences between the abstract and a textual summary derived solely from the video. That said, authors often present the abstract as a summary of the video, as it conveys the key contributions, objectives, and findings of the research, which are typically central to the content discussed.

\paragraph{Model} We have tested the plan-based approach on the video-based, audio-based, and text-based large models in our experiments. Our work does not aim to prove that the plan-based method is effective in all models of different modalities. Moreover, plan-based methods can take many different forms, and our work does not aim to identify the optimal planning approach for our dataset.

\paragraph{Scope} Our study focuses on video-to-text summarization within scientific domains. We have not investigated applying the plan-based method to other natural language processing (NLP) tasks, such as multimodal machine translation, multimodal question answering, or multimodal reasoning. Although the plan-based approach could likely be adapted to these tasks with minimal effort, such possibilities remain unexplored and warrant future investigation.

\paragraph{Automated Evaluation} While we employ a suite of automated metrics and hallucination detection methods to assess model performance on the test set, these metrics have inherent limitations and may fail to capture all aspects of model quality.

\paragraph{Human Evaluation} Similar to many earlier studies \cite{papalampidi-lapata-2023-hierarchical3d, krubinski-pecina-2023-mlask, krubinski-pecina-2024-towards, patil-etal-2024-refinesumm}, we only evaluate 50 video-summary pairs, a subset that may not represent the entire dataset. Additionally, while all evaluators are graduate students, they are not necessarily experts in video-to-text summarization and possess varying levels of reading and assessment skills. Consequently, although their evaluations are valuable, they should not be treated as the only indicator of performance.

\paragraph{LMM-as-Judge} Although the LMM-based judge paradigm enables large-scale and relatively consistent evaluations, it may inherit biases from its pretraining data, and its black-box nature makes the rating process difficult to interpret. Data contamination also remains a concern if \texttt{GPT-o1} is trained on overlapping data. We validate \texttt{GPT-o1}’s ratings with human evaluations on a small subset of samples, but this may not fully capture the model’s reliability across diverse topics, domains, or summary styles. Therefore, results should be interpreted with caution and supplemented by human judgment where possible.

\section*{Acknowledgements}
This project has received funding from the European Research Council (ERC) under the European Union’s Horizon 2020 Research and Innovation Programme (Grant Agreement No. 948878). Lapata acknowledges the support of the UK Engineering and Physical Sciences Research Council (Grant EP/W002876/1).

\begin{figure}[H] 
\centering
\includegraphics[width=0.9\columnwidth]{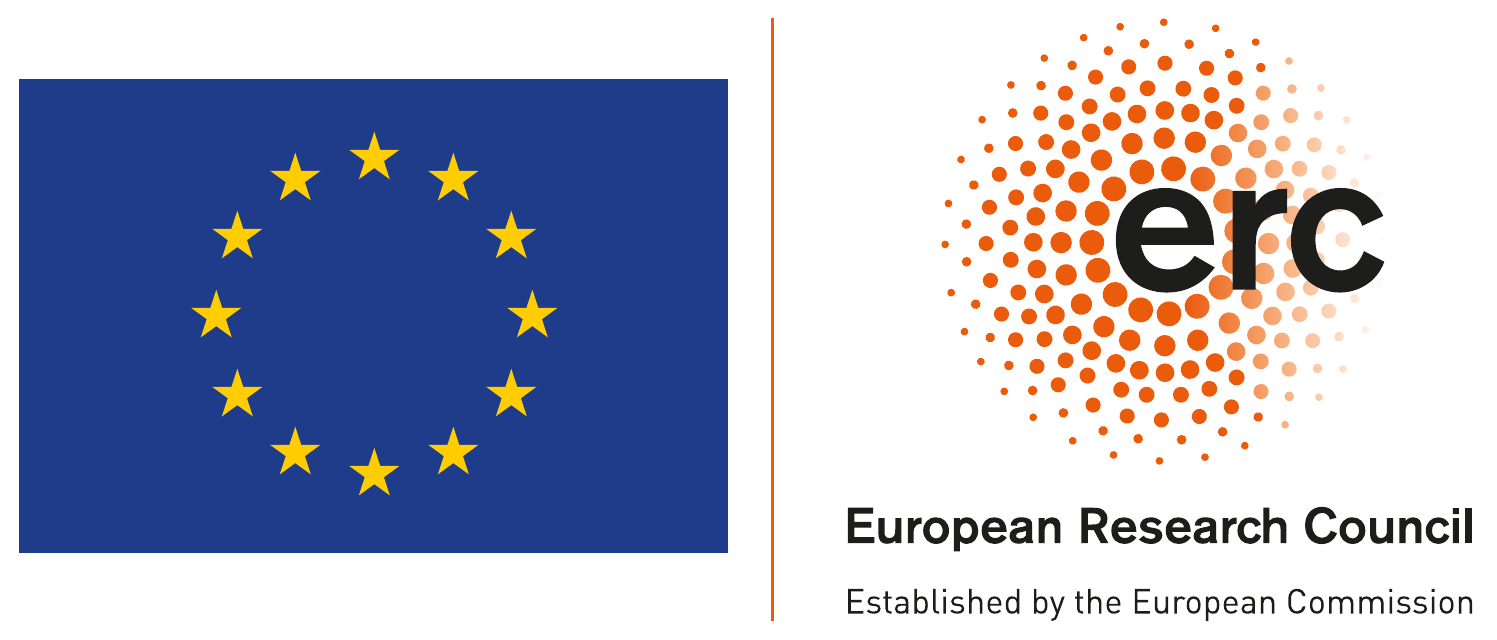}
\end{figure}

\bibliography{custom}

\appendix

\section{Copyright}
\label{copyrights}
According to the statement displayed on the \href{https://aclanthology.org/}{ACL Anthology} website, \textit{``Permission is granted to make copies for the purposes of teaching and research''}, allowing us to use the corresponding data. For \href{https://icml.cc/}{ICML} and \href{https://neurips.cc/}{NeurIPS}, we (the authors) have obtained written confirmation granting permission to use the paper titles, author lists, paper abstracts, full papers, and presentation videos available on their websites for research purposes.

\section{Quality Control}
\label{quality_control}
\paragraph{Manual Control}
We randomly select 500 video-summary pairs to assess whether the summaries provide accurate descriptions of the videos. Two Ph.D. candidates in Computer Science or Computational Linguistics perform binary judgments on these pairs. Across all 500 samples, neither evaluator rejected any sample.

\paragraph{Automated Control}
To go beyond the limited scope of manual checks, we employ \texttt{GPT-o1} for automated assessment using the same binary criteria across all data samples. The model initially flagged 39 pairs as potentially invalid. These flags were likely caused by difficulties in interpreting domain-specific terms or rare expressions and sensitivity to variations in summary length. After further manual review, all 39 samples were confirmed as valid and retained in the dataset.

\section{Data Sample}
\label{data_sample}

\begin{figure*}[t]
\centering
\begin{tcolorbox}[colback=gray!10!white, colframe=black!50!black, width=\textwidth, arc=0mm, boxrule=0.5mm, sharp corners=south]
   \begin{minipage}[t]{\textwidth}
       % \centering
     \includegraphics[width=\textwidth]{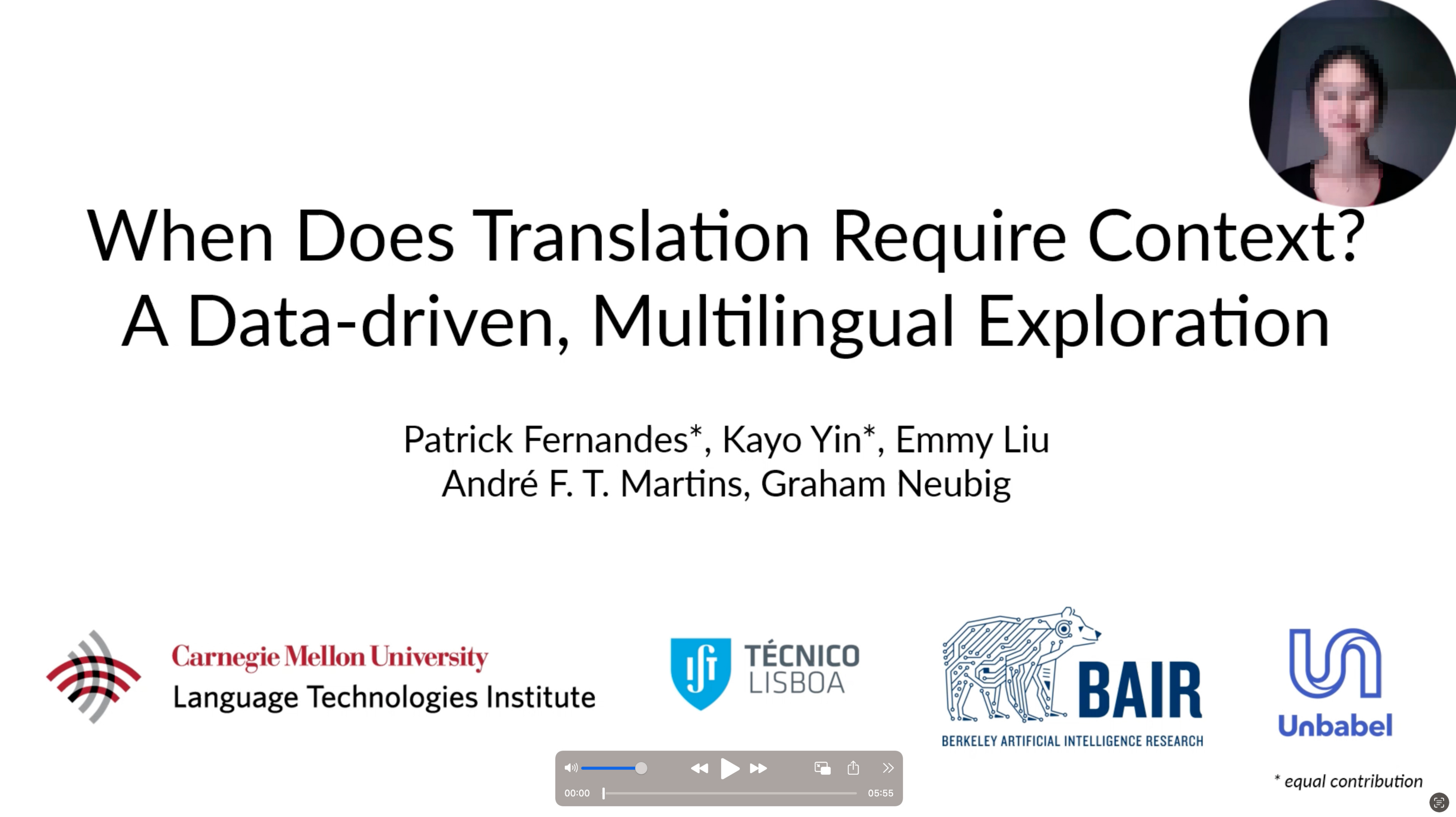}
    \end{minipage}
    \hfill
    \begin{minipage}[t]{\textwidth}
        \raggedright
        \begin{footnotesize}
        Although proper handling of discourse significantly contributes to the quality of machine translation (MT), these improvements are not adequately measured in common translation quality metrics. Recent works in context-aware MT attempt to target a small set of discourse phenomena during evaluation, however not in a fully systematic way. In this paper, we develop the Multilingual Discourse-Aware (MuDA) benchmark, a series of taggers that identify and evaluate model performance on discourse phenomena in any given dataset. The choice of phenomena is inspired by a novel methodology to systematically identify translations that require context. This methodology confirms the difficulty of previously studied phenomena while uncovering others which were not previously addressed. We find that commonly studied context-aware MT models make only marginal improvements over context-agnostic models, which suggests these models do not handle these ambiguities effectively. We release code and data for 14 language pairs to encourage the MT community to focus on accurately capturing discourse phenomena.
        \end{footnotesize}
    \end{minipage}
\end{tcolorbox}
\caption{A random sample from the VISTA dataset, originating from \citet{fernandes-etal-2023-translation}.}
\label{data_sample_1}
\end{figure*}

The VISTA dataset contains carefully curated video-text pairs, predominantly sourced from published papers, aiming to ensure a high standard of quality and relevance. The accompanying texts are designed to function as summaries of their respective videos, offering a concise representation of their content (see \autoref{data_sample_1}). Additionally, our dataset focuses on topics within the field of artificial intelligence, making it a good resource for research in AI-related video-to-text summarization and comprehension.

\section{Model Version Details}
\label{model_versions}
 
\autoref{tab:model_version} provides the detailed version identifiers for the models evaluated in our study, showing both model names as referenced in the main text and the specific versions used in our experiments.

\begin{table*}[t]
\centering
\scalebox{0.95}{
\tabcolsep=8pt
\begin{tabular}{l l c}
\toprule
Model & Version & Model Size\\
\midrule
\texttt{GPT-o1} \cite{achiam2023gpt} & \texttt{o1-2024-12-17} & Unknown\\
\texttt{Gemini 2.0} \cite{team2023gemini} & \texttt{Gemini 2.0 Flash} & Unknown\\
\texttt{Claude 3.5 Sonnet} \cite{anthropic_claude3_5_sonnet} & \texttt{claude-3-5-sonnet-20241022} & Unknown\\
% \hdashline
\texttt{LLaMA-3.1} \cite{touvron2023llama} & \texttt{LLaMA-3.1-8B-Instruct} & 8B \\
\texttt{Qwen2-Audio} \cite{chu2024qwen2} & \texttt{Qwen2-Audio-7B-Instruct} & 7B \\
% \hdashline
\texttt{Video-LLaMA} \cite{zhang-etal-2023-video}& \texttt{VideoLLaMA2-7B-16F} & 7B\\
\texttt{Video-ChatGPT}  \cite{maaz-etal-2024-video}& \texttt{Video-ChatGPT-7B} & 7B\\
\texttt{Video-LLaVA}  \cite{lin-etal-2024-video}& \texttt{Video-LLaVA-7B-hf} & 7B\\
\texttt{LLaMA-VID}  \cite{li2025llama}& \texttt{LLaMA-VID-7B-Full-224-Long-Video} & 7B\\
\texttt{LLaVA-NeXT-Interleave}  \cite{li2024llava}& \texttt{LLaVA-NeXT-Interleave-Qwen-7B} & 7B\\
\texttt{mPLUG-Owl3} \cite{ye2024mplug}& \texttt{mPLUG-Owl3-7B-241101}& 7B\\
\bottomrule
\end{tabular}
}
\caption{Model version details.}
\label{tab:model_version}
\end{table*}

\section{Hyper-parameters Settings}
\label{hyper-parameters_settings}
For all fine-tuning experiments, we utilize the AdamW optimizer \cite{loshchilovdecoupled} with $\beta_1$ = 0.9, $\beta_2$ = 0.999, $\epsilon$ = $10^{-9}$, and a weight decay of~0.1, combined with a warm-up ratio of~0.15. The initial learning rate is set to 5e-5, with cosine learning rate scheduling. DeepSpeed is configured with ZeRO-3 Offload. We set the random seed to 2025 and apply a dropout rate of 0.1. In the QLoRA setting, the rank~$r$ is set to~32, the scaling factor~$\alpha$ is set to 64, and the dropout rate for the low-rank matrices is~0.1. All other parameters follow the default settings of the \texttt{Transformers} library. 

During training, we save the checkpoint with the highest Rouge-2 F1 score on the validation set as the final model. All experiments are conducted over 16 epochs with a batch size of 16 and early stopping  (all models converged before 16 epochs). For model inference (including zero-shot learning), we employ a beam search with a beam of size~4, a length penalty of~3.0, a no-repeat n-gram size of 3, and the maximum number of new tokens generated is limited to 256. For video-based LMMs, the sampling rate is set to 0.1 fps, and the number of extracted frames is set to 32.

For closed-source models, results are obtained via API requests during the experimental period from 01/09/2024 to 10/02/2025. The hyperparameter settings for these API requests include a temperature of 1, top\_p of 1, a frequency penalty of 0.2, and a presence penalty of 0.2. All other parameters adhere to the default settings specified by their respective platforms.

\section{Automatic Evaluation Metrics}
\label{automatic_evaluation_metrics}

In line with common practice in video-to-text summarization research, we evaluate the model-generated summaries using the following metrics:
\begin{itemize}[leftmargin=8pt,itemsep=1pt,topsep=1pt,parsep=1pt]
\item ROUGE \cite{lin-2004-rouge}: measures n-gram overlap between machine-generated and human reference texts. We report F1 scores for Rouge-1 (R1), Rouge-2 (R2), and Rouge-Lsum (RLSUM).
\item SacreBLEU \cite{post-2018-call}: assesses linguistic consistency and fluency between generated and reference texts.
\item METEOR \cite{banerjee-lavie-2005-meteor}: calculates the harmonic mean of unigram precision and recall, placing greater emphasis on recall for a balanced evaluation.
\item BERTScore \cite{zhang2019bertscore}: uses contextual embeddings from BERT to evaluate semantic similarity between texts.
\item CIDEr-D \cite{vedantam2015cider}: evaluates the consensus between generated summaries and references by using TF-IDF weighting combined with a decay factor to reduce the impact of repeated terms.
\item VideoScore \cite{he-etal-2024-videoscore}: focuses on text-to-video alignment, evaluating how accurately video content matches the given text prompts using fine-grained multi-aspect scoring.
\item FactVC \cite{liu-wan-2023-models}: calculates the factual consistency of text with video content by aligning coarse-grained video-text similarity and precision-based fine-grained matching. The original values of FactVC range from 0 to 1, and in our experiments, we scale them by 100 to convert them into percentages.
\end{itemize}

\section{Plans Quality Validation}
\label{plan_quality_validation}
To validate the quality of the silver-standard plans generated by \texttt{GPT-o1}, we conduct a manual evaluation on 100 randomly selected samples. The evaluation is carried out by the same annotators involved in our human evaluation setup. Each annotator is asked to make a binary judgment on whether the generated plan question satisfied two validity criteria: (1) Local Coherence: The question is well-formed and semantically related to the summary; and (2) QUD-Alignment: Each sentence in the summary could plausibly serve as an answer to the question, consistent with the QUD framework.

We observe strong inter-annotator agreement (Fleiss’ $\kappa = 0.853$), indicating a high degree of consistency in decisions. In addition to this, we perform a manual error analysis to screen for systematic biases or recurrent flaws, such as overly generic phrasing, hallucinated entities, or structural redundancy. No such patterns are observed.

\section{Impact of Video Context on Summary Generation}
\label{video_context_impact}
We examine the impact of different video context configurations on summary generation, comparing   \texttt{mPLUG-Owl3} with \PlanSum. Unlike earlier experiments that use the full video as input, here only the first or last 10\% or 30\% of the video is provided as input.  We report results in the full fine-tuning setting. 

\begin{table}[h]
\centering
\scalebox{0.65}{
\tabcolsep=4.5pt
\begin{threeparttable}
\begin{tabular}{c l c c c c}
\toprule
Context & Model & R2 & RLsum & VideoScore & FactVC\\
\hline
\multirow{2}*{\rotatebox[origin=c]{360}{\mbox{All}}} 
& \texttt{mPLUG-Owl3} & 13.62 & 32.91 & 3.28 & 71.94 \\
~ & \PlanSum & 13.74 & 33.25 & 3.33 & 75.41 \\
\midrule
\multirow{2}*{\rotatebox[origin=c]{360}{\mbox{First 10\%}}} 
& \texttt{mPLUG-Owl3} & 6.31 & 25.44 & 2.37 & 51.02 \\
~ & \PlanSum & 7.37 & 27.38 & 2.52 & 52.39 \\
\midrule
\multirow{2}*{\rotatebox[origin=c]{360}{\mbox{First 30\%}}} 
& \texttt{mPLUG-Owl3} & 9.42 & 28.88 & 2.78 & 54.10 \\
~ & \PlanSum & 10.59 & 30.13 & 2.78 & 55.37 \\
\midrule
\multirow{2}*{\rotatebox[origin=c]{360}{\mbox{Last 10\%}}}
& \texttt{mPLUG-Owl3} & 6.53 & 27.34 & 2.51 & 53.64  \\
~ & \PlanSum & 7.62 & 29.73 & 2.77 & 55.93  \\
\midrule
\multirow{2}*{\rotatebox[origin=c]{360}{\mbox{Last 30\%}}} 
& \texttt{mPLUG-Owl3} & 7.32 & 29.17 & 2.82 & 57.36 \\
~ & \PlanSum & 10.72 & 31.29 & 2.98 & 62.05 \\
\bottomrule
\end{tabular}
\end{threeparttable}
}
\caption{Model performance under different video context configurations (full fine-tuning). The video content at the end is more helpful for summary generation.}
\label{tab:context_configurations}
\end{table}

The results in \autoref{tab:context_configurations} indicate that partial video context consistently underperforms compared to using the full video. Using the last part of the video generally produces better results than using the first part, as concluding sections often summarize key findings while opening sections primarily introduce background information. Additionally, utilizing 30\% of the video outperforms using only 10\%, highlighting that more content generally yields better outputs. Across all configurations, the \PlanSum model consistently outperforms \texttt{mPLUG-Owl3}.

\section{Impact of Text Context on Plan Generation}
\label{text_context_impact}
The generation of plan questions in our experiments is influenced by the target sentence and its context.  In our main experiments, plan questions are generated based on the target sentence and its preceding summary text (\texttt{Previous-Context}), in line with the original Question Under Discussion (QUD) requirements \cite{wu-etal-2023-qudeval, wu-etal-2023-elaborative, liu2025explanatorysummarizationdiscoursedrivenplanning}. We now assess configurations that generate questions only based on the target sentence (\texttt{No-Context}) or the entire summary (\texttt{All-Context}).

\begin{figure}[hbt]
  \centering 
  \includegraphics[width=0.49\textwidth]{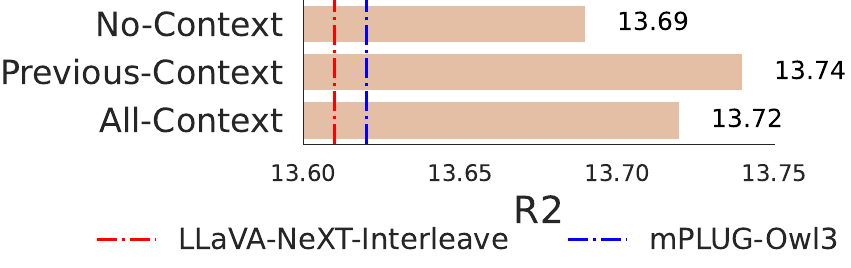}
  \caption{Impact of text context for plan generation.}
  \label{fig:text_contextual_variations_impact}
\end{figure}

As shown in \autoref{fig:text_contextual_variations_impact}, performance differences between different context configurations are relatively small (yet superior to models without planning components shown as red and blue dashed lines). \texttt{No-Context} shows the lowest performance but is the most cost-effective, as it requires the shortest input length for \texttt{GPT-o1} during question generation. \texttt{All-Context} achieves slightly better results but at the highest computational cost due to the long input length. \texttt{Previous-Context} is aligned with QUD and strikes a good balance, achieving the best performance for a moderate cost.

\section{Controllable Generation}
\label{controllable_generation}
An advantage of plan-based models is their ability to control the output summaries by modifying the plans used for generation. We investigate how modifying the structure and composition of these plans impacts the generated summaries, specifically comparing their performance against direct summary generation control through instructions. To this end, we design two controlled experiments:

\begin{itemize}[leftmargin=8pt,itemsep=1pt,topsep=1pt,parsep=1pt]
\item \textit{Summary Readability}: How question complexity affects readability, tailored for lay readers or expert readers.
\item \textit{Summary Length}: How the number of questions influences summary length, by removing 10\%, 30\%, and 60\% of questions.
\end{itemize}

\begin{table}[t]
\centering
\scalebox{0.9}{
\tabcolsep=5pt
\begin{tabular}{l c c c c}
\toprule
\multirow{2}{*}{Condition} & \multicolumn{2}{c}{\PlanSum} & \multicolumn{2}{c}{\texttt{GPT-o1}} \\
\cmidrule(lr){2-3} \cmidrule(l){4-5}
 & R2 & FRE & R2 & FRE \\
\midrule
No change        & 13.74 & 30.62 & 5.69 & 26.37 \\
Lay questions    & 13.38 & 35.17 & 4.26 & 28.94 \\
Expert questions & 13.24 & 23.54 & 4.13 & 24.33 \\
\bottomrule
\end{tabular}
}
\caption{Control experiment for summary readability. FRE $=$ Flesch Reading Ease.}
\label{tab:readability_control}
\end{table}

\begin{table}[t]
\centering
\scalebox{0.85}{
\tabcolsep=3pt
\begin{tabular}{l c c c c}
\toprule
\multirow{2}{*}{Condition} & \multicolumn{2}{c}{\PlanSum} & \multicolumn{2}{c}{\texttt{GPT-o1}} \\
\cmidrule(lr){2-3} \cmidrule(l){4-5}
 & R2 & Avg. \#Tokens & R2 & Avg. \#Tokens \\
\midrule
No deletion  & 13.74 & 202.39 & 5.69 & 267.32 \\
Delete 10\%  & 11.05 & 178.47 & 4.32 & 220.49 \\
Delete 30\%  & 10.41 & 137.72 & 3.17 & 192.42 \\
Delete 60\%  & 8.01 & 100.32 & 2.98 & 185.28 \\
\bottomrule
\end{tabular}
}
\caption{Control experiment for summary length.}
\label{tab:length_control}
\vspace{-10pt}
\end{table}

We note that the plan-based method employs an explicit planning component where each sentence is guided by a corresponding question that facilitates fine-grained control over the summary’s style or content. Specifically, after \texttt{PG} produces the plan, we use \texttt{GPT-o1} to edit it and then feed the edited questions back to \texttt{SG} for the final output. For \texttt{GPT-o1}, which operates in a zero-shot manner, we prepend constraints directly in the prompt. Specifically, \texttt{GPT-o1} generates an initial summary in one pass and then applies additional prompt-based instructions during a secondary rewriting step to control the output. Both control experiments (\autoref{tab:readability_control} and \autoref{tab:length_control}) reveal similar trends: While performance declines for both models, the plan-based method is more robust and controllable.

In the readability control experiment (\autoref{tab:readability_control}), both models show reductions in R2, but \PlanSum declines less, averaging an R2 loss of~0.43 compared to~1.50 for \texttt{GPT-o1}. Furthermore, \PlanSum controls readability more effectively, achieving a higher Flesch Reading Ease (FRE) score\footnote{The FRE score, which ranges from 0 to 100, measures text readability, with higher scores indicating easier-to-read content, and lower scores reflecting greater complexity.} of~35.17 for lay questions, compared to 28.94~for \texttt{GPT-o1}, and a lower FRE score of 23.54 for expert questions.

In the length control experiment (\autoref{tab:length_control}), R2 scores decline as content is removed, but the plan-based model aligns more closely with target compression ratios, producing summaries averaging 100.32 tokens under 60\% deletion, while \texttt{GPT-o1} generates longer summaries (185.28 tokens).

\section{Case Study and Error Analysis}
\label{case_study}
For our case study, we randomly select a sample \cite{kubler2020learning} from the test split. The analysis in \autoref{tab:case_study}  reveals differences in summary quality across models and against the human-written text. Specifically, \texttt{GPT-o1} often produces concise summaries but at the cost of precision. For example, it incorrectly claims that ``data splitting helps control test thresholds,'' which is a hallucination --- while data splitting ensures a tractable null distribution, it does not explicitly control test thresholds. Furthermore, its summaries frequently oversimplify complex concepts, reducing the depth of explanations and omitting crucial distinctions, such as the role of dependency calibration in the proposed method. Similarly, \texttt{mPLUG-Owl3} introduces factual inaccuracies, such as stating that data splitting ``ensures a reliable null distribution.'' This phrasing misleadingly implies that reliability is an inherent property of data splitting, whereas the correct point is that it makes the null distribution tractable rather than necessarily more reliable.

\PlanSum is more factually accurate than the other models. It correctly captures the main idea of full-sample hyperparameter learning and testing without data splitting. However, it still introduces subtle distortions, such as falsely suggesting a ``trade-off'' between test power and tractability, which misrepresents the actual relationship. These inaccuracies, while less severe than those in \texttt{GPT-o1} and \texttt{mPLUG-Owl3}, highlight the model’s tendency to infer unstated causal links, leading to potential misinterpretations. Despite the relative strengths of \PlanSum, all generated summaries fall short of human-written text. The model-generated outputs consistently struggle with informativeness, coherence, and factual accuracy. These shortcomings underscore the ongoing challenge of improving automated summarization systems to better align with human standards in both accuracy and clarity.

Controlled generation experiments reveal that hallucination issues are further amplified when imposing constraints on readability and length. Under readability control (\autoref{tab:readability_control_1}), \texttt{GPT-o1} is more likely to introduce fabricated or misleading content when forced to generate more complex outputs. This occurs because it lacks an explicit mechanism to ensure factual consistency while adapting to varying readability demands. Rather than relying on implicit internal heuristics, \PlanSum has an explicit planning mechanism which makes it less likely to introduce unsupported claims. Planning provides an additional layer of control, helping the model maintain factual alignment even as readability demands change.  
A similar trend is observed in length control experiments (\autoref{tab:readability_control_2}). As the compression ratio increases, \texttt{GPT-o1} struggles to balance conciseness and informativeness, sometimes hallucinating missing details to compensate for omitted content. This suggests that purely instruction-based control (i.e.,~prompting the model to shorten outputs) does not effectively enforce content retention, leading to greater inconsistencies. In contrast, the plan allows \PlanSum to selectively retain essential elements, reducing the risk of generating misleading content; it can also avoid answering deleted questions, to a certain extent.

These findings reinforce the advantages of plan-based control over instruction-based prompting. While neither approach fully eliminates hallucinations, planning provides a structured mechanism to manage content selection, ensuring greater alignment with the input source compared to freeform generative adjustments.

\begin{table*}[t]
\centering
\scalebox{1}{\tabcolsep=3pt
\begin{threeparttable}
\begin{tabular}{|p{1.9cm}|p{13cm}|}
\hline
\scriptsize Model & \scriptsize Output\\
\hline
\scriptsize Reference & \scriptsize Modern large-scale kernel-based tests such as maximum mean discrepancy (MMD) and kernelized Stein discrepancy (KSD) optimize kernel hyperparameters on a held-out sample via data splitting to obtain the most powerful test statistics. While data splitting results in a tractable null distribution, it suffers from a reduction in test power due to a smaller test sample size. Inspired by the selective inference framework, we propose an approach that enables learning the hyperparameters and testing on the full sample without data splitting. Our approach can correctly calibrate the test in the presence of such dependency and yield a test threshold in closed form. At the same significance level, our approach's test power is empirically larger than that of the data-splitting approach, regardless of its split proportion. \\
\hline
\scriptsize \texttt{GPT-o1} & \scriptsize Modern kernel-based tests like MMD and KSD use data splitting to tune hyperparameters. This splitting \textcolor{red}{helps control test thresholds} but reduces power because it splits the data into smaller parts. We propose a new method to tune hyperparameters without splitting the data. Our method provides better test power than data splitting. \\
\hline
\scriptsize \texttt{mPLUG-Owl3} & \scriptsize  Large-scale kernel-based tests like maximum mean discrepancy (MMD) and kernelized Stein discrepancy (KSD) usually split the data to tune hyperparameters. While this approach ensures \textcolor{red}{a reliable null distribution}, it reduces test power due to smaller sample sizes. We propose a method that avoids data splitting, enabling full-sample hyperparameter learning and testing. This method improves test power at the same significance level compared to data splitting, with \textcolor{red}{a closed-form threshold for the test.} \\
\hline
\scriptsize Plan Questions & \scriptsize q1: How do kernel-based tests like MMD and KSD typically optimize hyperparameters? \newline
q2: What is the key idea behind the proposed approach? \newline  
q3: What are the advantages of the proposed method compared to data-splitting methods? \\
\hdashline
\scriptsize \PlanSum  & \scriptsize Modern kernel-based tests like maximum mean discrepancy (MMD) and kernelized Stein discrepancy (KSD) typically split the data to optimize hyperparameters, \textcolor{red}{trading off} test power for a tractable null distribution. Inspired by the selective inference framework, we propose a novel approach that eliminates the need for data splitting, allowing full-sample hyperparameter learning and testing. Our method addresses dependencies introduced by learning and provides a closed-form test threshold while achieving higher test power than data-splitting methods, regardless of split proportions. \\
\hline
\end{tabular}
\end{threeparttable}
}
\vspace*{-.2cm}
\caption{Comparative case study of model outputs. Hallucinations are highlighted in \textcolor{red}{red} color.}
\label{tab:case_study}
\end{table*}

\begin{table*}[h]
\centering
\scalebox{1}{\tabcolsep=3pt
\begin{threeparttable}
\begin{tabular}{|p{1.9cm}|p{13cm}|}
\hline
\scriptsize Model & \scriptsize Output \\
\hline
\scriptsize \texttt{GPT-o1} & \scriptsize Modern kernel-based tests, such as MMD and KSD, use data splitting to \textcolor{red}{adjust hyperparameters}. This data splitting ensures \textcolor{red}{proper test thresholds} but weakens test power because the data is divided into smaller portions. We propose a new method that avoids splitting the data and provides better test power. \\
\hline
\scriptsize Plan  Questions & \scriptsize q1: What do MMD and KSD do with data splitting? \newline  
q2: What is the main purpose of the proposed method? \newline  
q3: Why is the proposed method better than data splitting?\\
\hdashline
\scriptsize \PlanSum  & \scriptsize Kernel-based tests like maximum mean discrepancy (MMD) and kernelized Stein discrepancy (KSD) typically divide the data into subsets to \textcolor{red}{adjust hyperparameters}. This approach ensures a manageable null distribution but reduces test power. Inspired by selective inference, we propose a new method that avoids splitting the data. This method uses the entire dataset for hyperparameter tuning and testing, textcolor{red}{eliminating dependencies introduced by learning}. It also \textcolor{red}{provides a clear, closed-form test threshold} and achieves higher test power compared to data-splitting methods, regardless of the split proportions. \\
\hline
\end{tabular}
\end{threeparttable}
}
\vspace*{-.2cm}
\caption{Decreasing reading difficulty: plan modifications vs instruction adjustments. Hallucinations are in \textcolor{red}{red}.}
\label{tab:readability_control_1}
\end{table*}
\begin{table*}[h!]
\centering
\scalebox{1}{\tabcolsep=3pt
\begin{threeparttable}
\begin{tabular}{|p{2cm}|p{13cm}|}
\hline
\scriptsize Model & \scriptsize Output \\
\hline
\scriptsize \texttt{GPT-o1} & \scriptsize Kernel-based statistical methodologies, exemplified by maximum mean discrepancy (MMD) and kernelized Stein discrepancy (KSD), conventionally employ data partitioning mechanisms to facilitate hyperparameter optimization. While this stratagem \textcolor{red}{ensures a well-calibrated null distribution}, it inherently compromises the statistical power of the test due to data fragmentation. We advance a paradigm wherein \textcolor{red}{hyperparameter selection is conducted holistically}, circumventing data splitting and consequently preserving statistical power. \\
\hline
\scriptsize Plan Questions & \scriptsize q1: Through what operational principles do MMD and KSD navigate the interplay between hyperparameter tuning and null distribution calibration? \newline
q2: How does the selective inference framework inform the methodological innovations of the proposed approach? \newline
q3: What are the quantifiable and theoretical advantages of the proposed strategy in circumventing the limitations imposed by data-splitting conventions? \\
\hdashline
\scriptsize \PlanSum  & \scriptsize Modern kernel-based tests like maximum mean discrepancy (MMD) and kernelized Stein discrepancy (KSD) typically split the data to optimize hyperparameters, \textcolor{red}{trading off} test power for a tractable null distribution. Inspired by the selective inference framework, we propose a novel approach that eliminates the need for data splitting, allowing full-sample hyperparameter learning and testing. Our method addresses dependencies introduced by learning and provides a closed-form test threshold while achieving higher test power than data-splitting methods, regardless of split proportions. \\
\hline
\end{tabular}
\end{threeparttable}
}
\vspace*{-.2cm}
\caption{Increasing reading difficulty: plan modifications vs. instruction adjustments. Hallucinations are in \textcolor{red}{red}.}
\label{tab:readability_control_2}

\end{table*}
\begin{table*}[htb!]
\centering
\scalebox{1}{\tabcolsep=3pt
\begin{threeparttable}
\begin{tabular}{|p{3cm}|p{12cm}|}
\hline
\scriptsize Model & \scriptsize Output \\
\hline
\scriptsize \texttt{GPT-o1} & \scriptsize Kernel-based tests like MMD and KSD use data splitting for hyperparameter tuning, \textcolor{red}{which controls thresholds} but reduces power. We propose a method that avoids splitting, \textcolor{red}{improving test power}. \\
\hline
\scriptsize Plan Questions & \scriptsize q1: How do kernel-based tests like MMD and KSD typically optimize hyperparameters? \newline
q2: What are the advantages of the proposed method compared to data-splitting methods? \\
\hdashline
\scriptsize \PlanSum & \scriptsize Kernel-based tests, such as maximum mean discrepancy (MMD) and kernelized Stein discrepancy (KSD), rely on data splitting for hyperparameter tuning. Our method \textcolor{red}{delivers analytically derived test thresholds} and consistently outperforms data-splitting techniques in test power, regardless of how the data is divided. \\
\hline
\end{tabular}
\end{threeparttable}
}
\vspace*{-.2cm}
\caption{Deleting 30\% of plan questions vs. instruction adjustments. Hallucinations are highlighted in \textcolor{red}{red}.}
\label{tab:length_control_case}
\end{table*}

\section{LMM-as-Judge Evaluation}
\label{lmm-as-judge}

To facilitate large-scale comparisons of model outputs, we adopt a method inspired by LLM-as-Judge \cite{liusie-etal-2024-llm, liu2024aligning, zheng:ea:2024, liu2025explanatorysummarizationdiscoursedrivenplanning}, extending it to use a large multimodal model \cite{chen2024mllm}. The proposed LMM-based evaluator incorporates both textual and video modalities and assesses the same summary quality dimensions used in our human evaluation for trend-level comparisons. Specifically, we use \texttt{GPT-o1} as the evaluator, following the hyperparameter settings in \autoref{hyper-parameters_settings}. To minimize potential bias from prior queries, the conversation history is reset before each evaluation. The instructions for the LMM-as-Judge evaluation are provided in \autoref{gpto1_summary_evaluation}.

We validate the agreement between \texttt{GPT-o1} and human ratings by comparing its ratings with human evaluations on the same 50 samples from the VISTA test set. We calculate Fleiss' Kappa between \texttt{GPT-o1} and mean human ratings across the dimensions of Faithfulness (\(\kappa\)=0.732), Relevance (\(\kappa\)=0.803), Informativeness (\(\kappa\)=0.730), Conciseness (\(\kappa\)=0.792) and Coherence (\(\kappa\)=0.721) at instance level. These results indicate that human evaluators and \texttt{GPT-o1} achieve substantial levels of agreement across these dimensions. Following this, we expand the evaluation to include all samples in our test set.

\begin{figure}[t]
  \centering \includegraphics[width=0.48\textwidth]{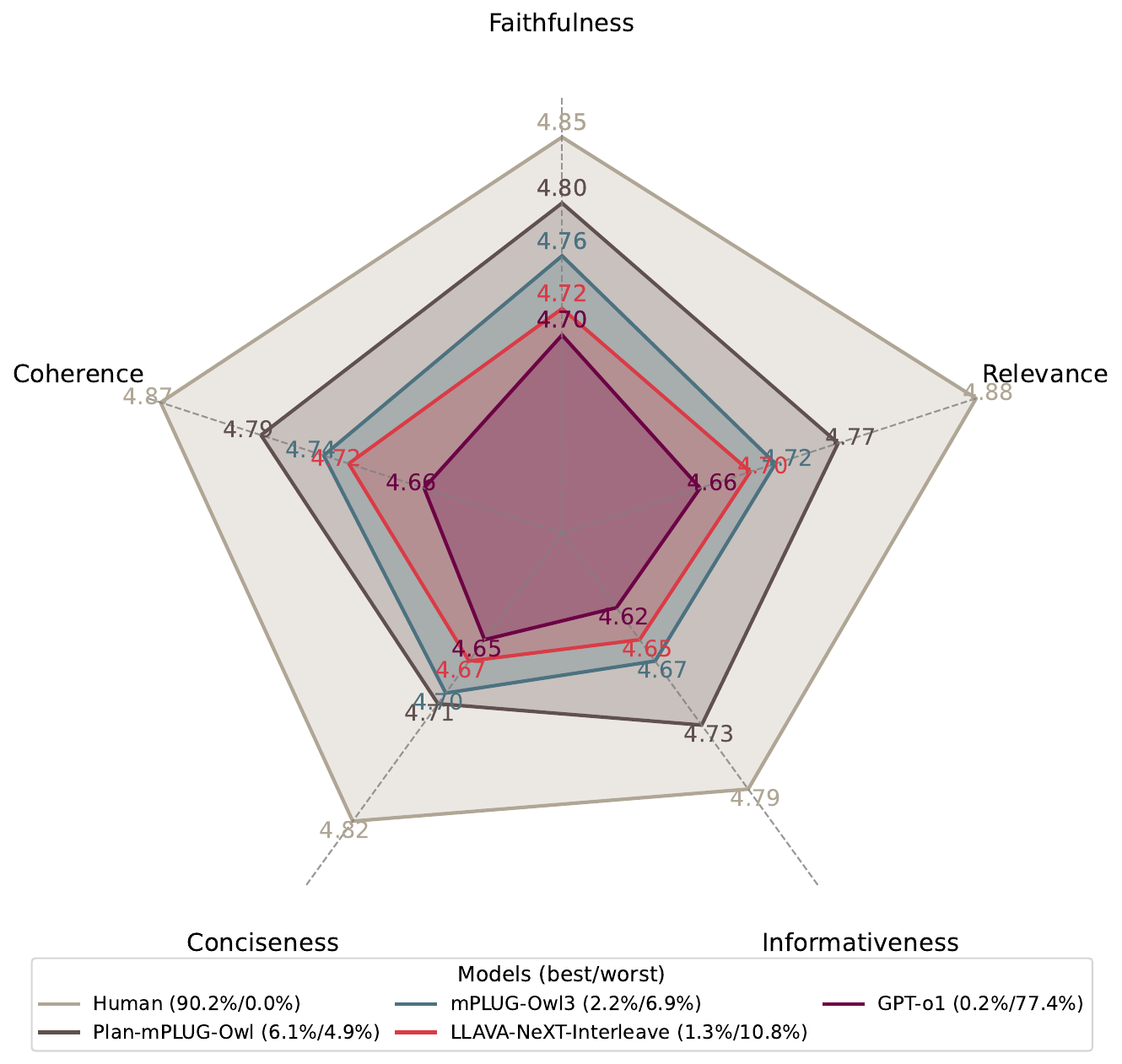}
  \caption{LMM-as-Judge evaluation results showing that human-written summaries consistently outperform neural models.}
  \label{fig:gpto1_evaluation}
\end{figure}

Compared to fine-tuned models, \texttt{GPT-o1} assigns the lowest scores to its own responses (see \autoref{fig:gpto1_evaluation}). Human-written summaries consistently receive the highest scores and are generally regarded as the best. Aligning with our human evaluations, \texttt{GPT-o1} also recognizes that the plan-based model outperformed other models. We further conduct paired t-tests to find that human summaries outperform all neural models across all metrics with statistical significance (\(p < 0.05\)). Moreover, the plan-based model demonstrates significantly better performance  (\(p < 0.05\)) than other neural models across all metrics except for conciseness. Our results also indicate that although the plan-based method can improve the performance of end-to-end models to some extent, there is still a gap between machine-generated and human summaries, which also reflects the challenging nature of our dataset.

\clearpage
%\onecolumn
\section{Prompts Used in Our Study}
\label{sec:prompts-study}

\begin{figure}[h]
\centering
\begin{tcolorbox}[
  colback=gray!10!white,
  colframe=black!50!black,
  title=Quality Control Guidelines,
  fonttitle=\bfseries,
  halign title=flush center,
]
\vspace*{-.2cm}
Evaluate each video-text pair to determine whether the text provides a concise and accurate summary of the corresponding video.
\begin{itemize}[leftmargin=8pt,itemsep=1pt,topsep=1pt,parsep=1pt]
    \item \textbf{Concise:} Ensure the text is brief, focused, and free of unnecessary details.
    \item \textbf{Accurate:} Verify that the text faithfully represents the video's content.
\end{itemize}
Make binary judgments (\texttt{Valid} or \texttt{Invalid}) for each pair. If flagged as \texttt{Invalid}, provide a brief justification.

\textbf{Answer:}\\
Judgment: (\texttt{Valid} or \texttt{Invalid})\\
Justification: (\texttt{Justification if flagged as invalid})
\vspace*{-.2cm}
\end{tcolorbox}
\vspace{-8pt}
\caption{Quality control guidelines.}
\label{fig:quality_control}
\end{figure}

\begin{figure}[h!]
\centering
\begin{tcolorbox}[colback=gray!10!white,colframe=black!50!black,title=Summary Generation (without plan), fonttitle=\bfseries, halign title=flush center]%, width=1\textwidth]
\vspace*{-.2cm}
Generate a summary for the provided content. \\
Content: \{\texttt{Video/Audio/Transcript/OCR}\} \\
Summary:
\vspace*{-.2cm}
\end{tcolorbox}
\vspace{-10pt}
\caption{Prompt to generate summaries without plans.}
\label{non_plan_based_summary_generation}
\end{figure}

\begin{figure}[h!]
\centering
\begin{tcolorbox}[
  colback=gray!10!white,
  colframe=black!50!black,
  title=Question  Generation,
  fonttitle=\bfseries,
  halign title=flush center,
]
\vspace*{-.2cm}
Generate a coherent and contextually relevant question based on the provided context and target sentence, ensuring that the target sentence can be treated as an answer to the generated question. \\
Context: \{\texttt{Context Text}\} \\
Target: \{\texttt{Target Sentence}\} \\
Question Sentence:
\vspace*{-.2cm}
\end{tcolorbox}
\vspace{-10pt}
\caption{Prompt for question generation.}
\label{plan_question_generation}
\end{figure}

\begin{figure}[h!]
\centering
\begin{tcolorbox}[colback=gray!10!white,colframe=black!50!black,title=Prompt for PG model, fonttitle=\bfseries, halign title=flush center]
\vspace*{-.2cm}
Generate a list of questions for the provided \{\texttt{Video/Audio/Transcript...}\}. \\
Content: \{\texttt{Video/Audio/Transcript...}\} \\
Questions:
\vspace*{-.2cm}
\end{tcolorbox}
\vspace{-10pt}
\caption{Prompt for PG model.}
\label{PG_model}
\end{figure}

\begin{figure}[h]
\centering
\begin{tcolorbox}[colback=gray!10!white,colframe=black!50!black,title=Prompt for SG model,fonttitle=\bfseries, halign title=flush center]%, width=1\textwidth]
\vspace*{-.2cm}
Generate a summary for the following \{\texttt{Video/Audio/Transcript...}\} based on the plan questions.\\
Content: \{\texttt{Video/Audio/Transcript...}\}. \\
Plan Questions: \{\texttt{Questions}\} \\
Ensure that the generated summary sequentially answers the plan questions.\\
Summary:
\vspace*{-.2cm}
\end{tcolorbox}
\vspace{-10pt}
\caption{Prompt for SG model.}
\label{SG_model}
\end{figure}

\begin{figure}[h]
\centering
\begin{tcolorbox}[colback=gray!10!white,colframe=black!50!black,title=Irrelevant Question Generation,fonttitle=\bfseries, halign title=flush center]%, width=1\textwidth]
\vspace*{-.2cm}
Randomly generate a question with a question mark. \\ 
Question Sentence: 
\vspace*{-.2cm}
\end{tcolorbox}
\vspace{-8pt}
\caption{Prompt used by GPT-o1 to generate irrelevant questions.}
\label{Irrelevant_Question_Generation}
\end{figure}

\begin{figure}[h]
\centering
\begin{tcolorbox}[colback=gray!10!white,colframe=black!50!black,title=Summary Readability Modification,fonttitle=\bfseries, halign title=flush center]%, width=1\textwidth]
\vspace*{-.2cm}
Rewrite the following text to further adjust the style or detail.

Here is the text to be rewritten: \{Text\}

Refine the above text to be more \{lay/expert\} style.\\
Modified Text:
\vspace*{-.2cm}
\end{tcolorbox}
\vspace{-8pt}
\caption{Summary readability modification.}
\label{fig:Summary_Readability}
\end{figure}

\begin{figure}[h]
\centering
\begin{tcolorbox}[
  colback=gray!10!white,
  colframe=black!50!black,
  title=Summary Length Modification,
  fonttitle=\bfseries,
  halign title=flush center,
 % width=1\textwidth
]
\vspace*{-.2cm}
Rewrite the following text to further adjust the style or detail.

Here is the text to be rewritten: \{Text\}

Shorten the above text by about \{10\% / 30\% / 60\%\}. Focus on the key points and remove less critical details.\\
Modified Text:
\vspace*{-.2cm}
\end{tcolorbox}
\vspace{-8pt}
\caption{Summary length modification.}
\label{fig:Summary_Length}
\end{figure}

\begin{figure}[h]
\centering
\begin{tcolorbox}[colback=gray!10!white,colframe=black!50!black,title=Plan Readability Modification,fonttitle=\bfseries, halign title=flush center]%, width=1\textwidth]
\vspace*{-.2cm}
Rewrite the following questions to further adjust the style or detail.

Here are the questions to be rewritten:\\ 
1. \{Q1\}\\
2. \{Q2\}\\
...

Refine the above questions to be more \{lay/expert\} style.\\
Modified Questions:
\vspace*{-.2cm}
\end{tcolorbox}
\vspace{-8pt}
\caption{Plan readability modification.}
\label{fig:Plan_Readability}
\end{figure}

\onecolumn
\section{Human Evaluation Guidelines}
\label{human_evaluation_guideline}
\begin{figure*}[h]
\footnotesize
\centering
\begin{tcolorbox}[]
\vspace*{-.2cm}
\paragraph{Prerequisites} To participate in this evaluation, you must meet the following two criteria: (1) be a Master's or Ph.D. student in Computer Science or Computational Linguistics, and (2) demonstrate English proficiency at C2 level or higher.\footnote{\url{https://en.wikipedia.org/wiki/C2_Proficiency}} If you do not meet both criteria, we kindly ask you to refrain from participating in this task. Eligible participants are encouraged to follow the instructions below carefully.\\

\paragraph{Instructions} The following section provides detailed descriptions of the evaluation metrics and criteria used in this study. Please review the accompanying source video and the candidate summaries thoroughly. After evaluating each summary, assign scores based on the five criteria below, using a 1-to-5 Likert scale where higher scores indicate better quality:

\begin{itemize}[leftmargin=8pt,itemsep=1pt,topsep=1pt,parsep=1pt]
    \item \textbf{Faithfulness:} Assess the accuracy of the summary in representing the content of the source video. A faithful summary should adhere closely to the source material, avoiding contradictions, misinterpretations, or unverified information.
    \item \textbf{Relevance:} Measure how well the summary includes the topics and themes central to the source video. A relevant summary should focus on the content that is most pertinent to the original video.
    \item \textbf{Informativeness:} Evaluate the extent to which the summary captures the main points and essential details of the source video. An informative summary should provide a clear and comprehensive understanding of the video’s core ideas and findings.
    \item \textbf{Conciseness:} Determine the efficiency of the summary in conveying information. A concise summary should avoid redundancy and extraneous details while retaining all critical information from the source video.
    \item \textbf{Coherence:} Examine the logical flow and overall structure of the summary. A coherent summary should present information in an organized and easy-to-follow manner, ensuring that ideas connect naturally and transitions between points are smooth.
\end{itemize} 

\paragraph{Rating System}
For each metric, use the following Likert scale:
\begin{itemize}[leftmargin=8pt,itemsep=1pt,topsep=1pt,parsep=1pt]
    \item 1 (Worst):  Does not meet the criteria at all.
    \item 2 (Poor): Meets the criteria minimally.
    \item 3 (Fair): Meets the criteria adequately.
    \item 4 (Good): Meets the criteria well.
    \item 5 (Best): Fully meets the criteria.
\end{itemize}

\paragraph{Overall Ranking}
After assigning scores to each summary for the individual criteria, rank all candidates from best to worst based on their overall quality. Consider the summaries' performance across all criteria when determining the final rankings.

\end{tcolorbox}
\caption{A snapshot of the experimental instructions provided to human evaluators.}
\end{figure*}

%\onecolumn
\section{Prompt for \texttt{GPT-o1} to Evaluate Summary Quality}
\label{gpto1_summary_evaluation}
\begin{figure*}[ht!]
\footnotesize
\centering
\begin{tcolorbox}
\vspace*{-.2cm}
\textbf{Source Video:} \{\texttt{Source Video}\} \\
\textbf{Candidate Summary:} \{\texttt{Candidate Summary}\} \\ 
You are tasked with evaluating the quality of the candidate summary based on the provided source video. Please adhere strictly to the following evaluation guidelines and scoring criteria to ensure a consistent and objective evaluation. 

\textbf{Evaluation Guidelines:} \{\texttt{Guidelines}\} 

\textbf{Instructions for Output:}
\begin{itemize}[leftmargin=8pt,itemsep=1pt,topsep=1pt,parsep=1pt]
    \item Provide your evaluation using the following format, outputting scores only.
    \item Assign a score from 1 to 5 for each dimension, with 1 being the lowest and 5 being the highest.
\end{itemize}

\textbf{Output Format:}
\begin{itemize}[leftmargin=8pt,itemsep=1pt,topsep=1pt,parsep=1pt]
    \item Faithfulness: [Score]
    \item Relevance: [Score]
    \item Informativeness: [Score]
    \item Conciseness: [Score]
    \item Coherence: [Score]
\end{itemize}

If you encounter ambiguity in evaluating any dimension, prioritize adherence to the evaluation guidelines and provide the most accurate score possible based on the provided information. Do not include any additional comments or justifications in your response.
\vspace*{-.2cm}
\end{tcolorbox}
\vspace*{-.2cm}
\caption{Prompt for \texttt{GPT-o1} to evaluate summary quality.}
\end{figure*}

\end{document}